\pdfoutput=1

\documentclass[11pt]{article}

\usepackage[final]{acl}

\usepackage{times}
\usepackage{latexsym}

\usepackage[T1]{fontenc}

\usepackage[utf8]{inputenc}

\usepackage{microtype}

\usepackage{inconsolata}

\usepackage{adjustbox}
\usepackage{amsmath}
\usepackage[amssymb]{SIunits}
\usepackage{amssymb}

\usepackage{algorithm}

\usepackage{supertabular}
\usepackage{array}
\usepackage{psfrag}
\usepackage{hhline}
\usepackage{type1cm}
\usepackage[normalem]{ulem}
\usepackage[noend]{algpseudocode}
\usepackage{pgfplots}
\pgfplotsset{compat=1.14}

\usepackage{ctable}
\usepackage{tikz}
\usetikzlibrary{positioning,arrows,trees,shapes,fit}
\usepackage{tikz-cd}
\usetikzlibrary{backgrounds}

\tikzset{
    >=stealth',
    punkt/.style={
           rectangle,
           rounded corners,
           draw=black, very thick,
           text width=6.5em,
           minimum height=2em,
           text centered},
    pil/.style={
           ->,
           thick,
           shorten <=2pt,
           shorten >=2pt,}
}

\usepackage{multicol}
\usepackage{multirow}
\usepackage{wrapfig}
\usepackage{enumerate}
\usepackage{enumitem}
\usepackage{subcaption}
\usepackage{mathtools}
\usepackage{lipsum}
\usepackage{paracol}
\usepackage{pifont}
\usepackage{verbatim}
\newcommand{\xmark}{\ding{55}}%
\usepackage[utf8]{inputenc}
\usepackage[T2A, T5, T1]{fontenc}

\usepackage{CJKutf8}
\usepackage{color,soul}               

\usepackage[main=english,russian,vietnamese]{babel}

\usepackage{pgfplotstable} 

\usepackage{graphicx}

\definecolor{LRed}{rgb}{1,.9,.9}
\definecolor{LGreen}{rgb}{.9,1,0.9} 
\definecolor{LBlue}{rgb}{.9,.9,1}
\definecolor{LYellow}{rgb}{1,1,0.9}

\definecolor{lightblue}{rgb}{0.68, 0.85, 0.9}
\definecolor{lavender}{rgb}{0.9, 0.9, 0.98}
\definecolor{lightyellow}{rgb}{1.0, 1.0, 0.88}
\definecolor{magicmint}{rgb}{0.67, 0.94, 0.82}
\definecolor{palepink}{rgb}{0.98, 0.85, 0.87}
\definecolor{bubbles}{rgb}{0.91, 1.0, 1.0}

\usepackage{float}

\usepackage{enumerate}

\newcommand{\red}[1]{\textcolor{red}{#1}}
\newcommand{\blue}[1]{\textcolor{blue}{#1}}

\newcommand{\purple}[1]{\textcolor{purple}{#1}}

\newcommand{\boxpink}[1]{\colorbox{pink}{#1}}
\newcommand{\boxyellow}[1]{\colorbox{yellow}{#1}}
\newcommand{\boxlime}[1]{\colorbox{lime}{#1}}
\newcommand{\boxlightyellow}[1]{\colorbox{lightyellow}{#1}}
\newcommand{\boxlightblue}[1]{\colorbox{lightblue}{#1}}

\definecolor{darkblue}{rgb}{0.0, 0.0, 0.5}
\definecolor{airforceblue}{rgb}{0.0, 0.0, 0.8}
\definecolor{crimson}{rgb}{0.8, 0.0, 0.0}

\newcommand{\eg}{\emph{e.g.,}}

\newcommand{\Ni}{({\em i})~}
\newcommand{\Nii}{({\em ii})~}
\newcommand{\Niii}{({\em iii})~}

\usepackage{amsmath,amsfonts,bm}

\def\eqref#1{equation~\ref{#1}}

\def\1{\bm{1}}


\DeclareMathAlphabet{\mathsfit}{\encodingdefault}{\sfdefault}{m}{sl}
\SetMathAlphabet{\mathsfit}{bold}{\encodingdefault}{\sfdefault}{bx}{n}

\pgfplotstableread{
id    lang  percent 
1       tu	0.00002
2       ki	0.00004
3       bm	0.00004
4       ak	0.00007
5       tg	0.00007
6       ss	0.00007
7       ny	 0.0001
8       tn	 0.0002
9       ln	 0.0002
10      ns	 0.0002
11      fo	 0.0002
12      rd	 0.0003
13      wo	 0.0004
14      lu	 0.0004
15      sn	  0.001
16      zu	  0.001
17      ig	  0.001
18      xh	  0.001
19      kw	  0.003
20      yo	  0.006
21      sw	   0.02
22      as	   0.01
23      or	   0.04
24      gu	   0.04
25      mr	   0.05
26      pa	   0.05
27      kn	   0.06
28      ne	   0.07
29      te	   0.09
30      ml	    0.1
31      ur	    0.1
32      ta	    0.2
33      bn	    0.5
34      hi	    0.7
}\langpercent

\pgfplotstableread{
id    lang  percent 
1       tu	0.00002
2       ki	0.00004
3       bm	0.00004
4       ak	0.00007
5       tg	0.00007
6       ss	0.00007
7       ny	 0.0001
8       tn	 0.0002
9       ln	 0.0002
10      ns	 0.0002
11      fo	 0.0002
12      rd	 0.0003
13      wo	 0.0004
14      lu	 0.0004
15      sn	  0.001
16      zu	  0.001
17      ig	  0.001
18      xh	  0.001
19      kw	  0.003
20      yo	  0.006
21      sw	   0.02
}\langpercentafrican

\pgfplotstableread{
id    lang  percent 
22      as	   0.01
23      or	   0.04
24      gu	   0.04
25      mr	   0.05
26      pa	   0.05
27      kn	   0.06
28      ne	   0.07
29      te	   0.09
30      ml	    0.1
31      ur	    0.1
32      ta	    0.2
33      bn	    0.5
34      hi	    0.7
}\langpercentindic

\pgfplotstableread{
1   53.0  36.027  8.219   51.0   99.726  1
2   56.0  21.918  4.110   53.0   12.329  1
3   56.0  16.438  6.849   55.0   39.726  1
4   64.0  2.740   2.740   56.0   28.767  1
5   55.0  1.370   6.849   52.0   1.3700  2
6   58.0  2.740   16.43   52.0   57.534  2
7   58.0  0.000   0.000   55.0   32.877  2
8   63.0  6.849   5.479   56.0   15.068  2
9   56.0  12.329  6.849   50.0   20.548  3
10  54.0  4.110   8.219   52.0   35.616  3
11  58.0  12.329  6.849   54.0   20.548  3
12  64.0  4.110   8.219   57.0   35.616  3
}\dataA

\pgfplotstableread{
id    lang  percent  sup_toen  sup_fromen  our_toen  our_fromen
1       as	    0.01	  46.21	      27.13	     46.78		  26.57
2       or	    0.04	  49.99	      32.00	     50.86		  31.24
3       gu	    0.04	  51.63	      38.23	     50.09		  37.63
4       mr	    0.05	  33.08	      25.73	     33.70		  24.97
5       pa	    0.05	  54.04	      31.99	     53.35		  32.27
6       kn	    0.06	  47.82	      34.17	     47.94		  33.48
7       ne	    0.07	  47.07	      35.91	     48.26		  35.76
8       te	    0.09	  44.44	      36.70	     44.85		  37.26
9       ml	     0.1	  38.67	      25.56	     35.47		  25.39
10      ur	     0.1	  50.20	      37.00	     52.36		  37.45
11      ta	     0.2	  43.93	      38.63	     46.47		  38.80
12      bn	     0.5	  52.82	      42.56	     53.25		  42.78
13      hi	     0.7	  55.18	      44.94	     55.71		  45.36
}\dataindic

\pgfplotstableread{
id    lang  percent  percent2  sup_toen  sup_fromen  our_toen  our_fromen
1       tu	0.00002     0.002	    26.43		   9.89	      26.39	    10.91
2       ki	0.00004     0.004	    25.26		  11.59	      20.51	    9.12
3       bm	0.00004     0.004	    20.69		   6.35	      21.73	    8.45
4       ak	0.00007     0.007	    25.97		  12.67	      25.15	    12.23
5       tg	0.00007     0.007	    24.69		  11.86	      25.67	    9.75
6       ss	0.00007     0.007	    24.53		  13.04	      21.61	    11.28
7       ny	 0.0001      0.01	    27.74		  12.85	      28.91	    12.96
8       tn	 0.0002      0.02	    24.69		  14.48	      26.38	    14.18
9       ln	 0.0002      0.02	    28.73		  17.92	      28.45	    16.95
10      ns	 0.0002      0.02	    26.42		  12.84	      27.23	    12.28
11      fo	 0.0002      0.02	    19.73		   5.12	      19.45	    4.84
12      rd	 0.0003      0.03	    35.16		  16.36	      34.54	    15.22
13      wo	 0.0004      0.04	    27.27		  14.14	      28.63	    13.61
14      lu	 0.0004      0.04	    26.30		  13.68	      27.35	    12.33
15      sn	  0.001       0.1	    27.44		  12.52	      27.68	    12.10
16      zu	  0.001       0.1	    23.45		   9.56	      21.98	    9.51
17      ig	  0.001       0.1	    32.63		  17.61	      33.87	    17.46
18      xh	  0.001       0.1	    25.89		  12.53	      26.56	    13.19
19      kw	  0.003       0.3	    40.10		  20.52	      40.14	    20.71
20      yo	  0.006       0.6	    31.78		  20.21	      34.46	    21.93
21      sw	   0.02         2	    56.51		  47.68	      56.42	    47.06
}\dataafrican

\pgfplotstableread{
id    lang  percent   sup_toen  sup_fromen  our_toen  our_fromen
1       tu	0.00002 	  26.43		 9.89		26.39		10.91
2       ki	0.00004 	  25.26		11.59		20.51		9.12
3       bm	0.00004 	  20.69		 6.35		21.73		8.45
4       ak	0.00007 	  25.97		12.67		25.15		12.23
5       tg	0.00007 	  24.69		11.86		25.67		9.75
6       ss	0.00007 	  24.53		13.04		21.61		11.28
7       ny	 0.0001 	  27.74		12.85		28.91		12.96
8       tn	 0.0002 	  24.69		14.48		26.38		14.18
9       ln	 0.0002 	  28.73		17.92		28.45		16.95
10      ns	 0.0002 	  26.42		12.84		27.23		12.28
11      fo	 0.0002 	  19.73		 5.12		19.45		4.84
12      rd	 0.0003 	  35.16		16.36		34.54		15.22
13      wo	 0.0004 	  27.27		14.14		28.63		13.61
14      lu	 0.0004 	  26.30		13.68		27.35		12.33
15      sn	  0.001 	  27.44		12.52		27.68		12.10
16      zu	  0.001 	  23.45		 9.56		21.98		9.51
17      ig	  0.001 	  32.63		17.61		33.87		17.46
18      xh	  0.001 	  25.89		12.53		26.56		13.19
19      kw	  0.003 	  40.10		20.52		40.14		20.71
20      yo	  0.006 	  31.78		20.21		34.46		21.93
21      sw	   0.02 	  56.51		47.68		56.42		47.06
22      as	   0.01	    46.21		 27.13	 46.78	26.57
23      or	   0.04	    49.99		 32.00	 50.86	31.24
24      gu	   0.04	    51.63		 38.23	 50.09	37.63
25      mr	   0.05	    33.08		 25.73	 33.70	24.97
26      pa	   0.05	    54.04		 31.99	 53.35	32.27
27      kn	   0.06	    47.82		 34.17	 47.94	33.48
28      ne	   0.07	    47.07		 35.91	 48.26	35.76
29      te	   0.09	    44.44		 36.70	 44.85	37.26
30      ml	    0.1	    38.67		 25.56	 35.47	25.39
31      ur	    0.1	    50.20		 37.00	 52.36	37.45
32      ta	    0.2	    43.93		 38.63	 46.47	38.80
33      bn	    0.5	    52.82		 42.56	 53.25	42.78
34      hi	    0.7	    55.18		 44.94	 55.71	45.36
}\dataafricanindic

\pgfplotstableread{
id    lang  percent   sup_toen  sup_fromen  our_toen  our_fromen	supfen_m_ourfen   bloom_tok_ratio     openai_tok_ratio    openai_sup_toen     openai_sup_fromen   openai_m_blo_toen	openai_m_blo_fromen
1       tu	0.00002 	26.43		  9.89	 26.39	10.91		-1.02                           2.210               2.802               29.62          	      14.75           3.19	        4.86
2       ki	0.00004 	25.26		 11.59	 20.51	9.12		 2.47                           2.527               3.504               25.10          	      13.45           -0.16	        1.86
3       bm	0.00004 	20.69		  6.35	 21.73	8.45		 -2.1                           1.917               2.687               25.10          	      11.69           4.41	        5.34
4       ak	0.00007 	25.97		 12.67	 25.15	12.23		 0.44                           2.067               2.803               26.11          	      24.56           0.14	        11.89
5       tg	0.00007 	24.69		 11.86	 25.67	9.75		 2.11                           2.026               2.471               33.95          	      22.88           9.26	        11.02
6       ss	0.00007 	24.53		 13.04	 21.61	11.28		 1.76                           1.984               2.365               25.10          	      12.93           0.57	        -0.11
7       ny	 0.0001 	27.74		 12.85	 28.91	12.96		-0.11                           1.810               2.295               30.29          	      20.50           2.55	        7.65
8       tn	 0.0002 	24.69		 14.48	 26.38	14.18		  0.3                           2.063               2.434               26.12          	      20.13           1.43	        5.65
9       ln	 0.0002 	28.73		 17.92	 28.45	16.95		 0.97                           1.662               2.042               30.42          	      23.92           1.69	        6
10      ns	 0.0002 	26.42		 12.84	 27.23	12.28		 0.56                           1.960               2.343               28.56          	      21.70           2.14	        8.86
11      fo	 0.0002 	19.73		  5.12	 19.45	4.84		 0.28                           2.220               4.100               24.61          	      8.260           4.88	        3.14
12      rd	 0.0003 	35.16		 16.36	 34.54	15.22		 1.14                           1.653               2.337               30.15          	      17.12           -5.01	        0.76
13      wo	 0.0004 	27.27		 14.14	 28.63	13.61		 0.53                           1.690               2.154               28.84          	      19.77           1.57	        5.63
14      lu	 0.0004 	26.30		 13.68	 27.35	12.33		 1.35                           1.690               2.191               28.84          	      19.77           2.54	        6.09
15      sn	  0.001 	27.44		 12.52	 27.68	12.10		 0.42                           1.826               2.322               35.82          	      24.23           8.38	        11.71
16      zu	  0.001 	23.45		  9.56	 21.98	9.51		 0.05                           1.761               2.416               27.12          	      18.40           3.67	        8.84
17      ig	  0.001 	32.63		 17.61	 33.87	17.46		 0.15                           1.703               3.397               29.10          	      19.12           -3.53	        1.51
18      xh	  0.001 	25.89		 12.53	 26.56	13.19		-0.66                           1.691               2.298               25.00          	      14.73           -0.89	        2.2
19      kw	  0.003 	40.10		 20.52	 40.14	20.71		-0.19                           1.602               2.398               30.11          	      15.77           -9.99	        -4.75
20      yo	  0.006 	31.78		 20.21	 34.46	21.93		-1.72                           1.671               3.899               29.11          	      16.71           -2.67	        -3.5
21      sw	   0.02 	56.51		 47.68	 56.42	47.06		 0.62                           1.251               2.147               51.08          	      43.17           -5.43	        -4.51
22      as	   0.01	    46.21		 27.13	 46.78	26.57		 0.56                           1.439               9.878               33.96          	      19.06           -12.25	    -8.07
23      or	   0.04	    49.99		 32.00	 50.86	31.24		 0.76                           1.378               13.47               37.81          	      16.67           -12.18	    -15.33
24      gu	   0.04	    51.63		 38.23	 50.09	37.63		  0.6                           1.369               12.30               40.79          	      23.22           -10.84	    -15.01
25      mr	   0.05	    33.08		 25.73	 33.70	24.97		 0.76                           1.230               7.897               26.12          	      28.53           -6.96	        2.8
26      pa	   0.05	    54.04		 31.99	 53.35	32.27		-0.28                           1.453               7.932               40.09          	      26.38           -13.95	    -5.61
27      kn	   0.06	    47.82		 34.17	 47.94	33.48		 0.69                           1.345               13.75               36.40          	      20.01           -11.42	    -14.16
28      ne	   0.07	    47.07		 35.91	 48.26	35.76		 0.15                           1.196               7.679               36.54          	      25.34           -10.53	    -10.57
29      te	   0.09	    44.44		 36.70	 44.85	37.26		-0.56                           1.361               13.17               27.88          	      24.66           -16.56	    -12.04
30      ml	    0.1	    38.67		 25.56	 35.47	25.39		 0.17                           1.393               15.32               26.84          	      15.11           -11.83	    -10.45
31      ur	    0.1	    50.20		 37.00	 52.36	37.45		-0.45                           1.386               6.348               45.61          	      34.56           -4.59	        -2.44
32      ta	    0.2	    43.93		 38.63	 46.47	38.80		-0.17                           1.298               15.64               29.10          	      18.84           -14.83	    -19.79
33      bn	    0.5	    52.82		 42.56	 53.25	42.78		-0.22                           1.190               9.723               47.11          	      29.67           -5.71	        -12.89
34      hi	    0.7	    55.18		 44.94	 55.71	45.36		-0.42                           1.308               7.520               53.61          	      39.62           -1.57	        -5.32
}\dataafricanindic

\pgfplotstableread{

id  lorar		params_n    indic2en    en2indic  unsup_indic2en unsup_en2indic x2en_diff   en2x_diff
1		8	      $3e^7$	41.19		23.98		    39.88		24.41       1.31	-0.43
2		16	      $5e^7$	41.25		26.01		    39.88		24.41       1.37	1.6
3		32	      $1e^8$	41.53		27.31		    39.88		24.41       1.65	2.9
4		64	      $2e^8$	41.59		28.69		    39.88		24.41       1.71	4.28
5		128	      $5e^8$	41.71		29.69		    39.88		24.41       1.83	5.28
6		256	      $1e^9$	41.89		31.51		    39.88		24.41       2.01	7.1
7		512	      $2e^9$	42.19		32.72		    39.88		24.41       2.31	8.31
}\datafinetuneindic

\usepackage{hyperref}
\usepackage{cleveref}
\usepackage{url}

\usepackage{microtype}

\title{Democratizing LLMs for Low-Resource Languages by Leveraging their English Dominant Abilities with Linguistically-Diverse Prompts}

\author{
    Xuan-Phi Nguyen$^1$, Sharifah Mahani Aljunied$^1$, Shafiq Joty$^2$,  Lidong Bing$^1$\\
    $^1$DAMO Academy, Alibaba Group\\
    $^2$Nanyang Technological University, Singapore\\
   \small{\texttt{nxphi47@gmail.com,  mahani.aljunied@alibaba-inc.com,  srjoty@ntu.edu.sg,  l.bing@alibaba-inc.com} }
}

\begin{document}
\maketitle
\begin{abstract}
Large language models (LLMs) are known to perform tasks by simply observing few exemplars. Moreover, competent generative capabilities of LLMs are observed mostly in high-resource languages, while their performances among under-represented languages fall behind due to pre-training data imbalance. To elicit LLMs' ability onto low-resource languages without any supervised data, we propose to assemble synthetic exemplars from a diverse set of high-resource languages. These prompts can directly induce generative capabilities in low-resource languages and serve as intra-lingual exemplars to even improve tasks in these languages. Our unsupervised prompting method performs on par with supervised few-shot learning in LLMs of different sizes for translations between English and 34 Indic and African languages, and surpasses supervised prompting in non-English tasks. The method also significantly improves low-resource performances in many other intra-lingual tasks like summarization (XLSum), question answering (XQUAD \& TydiQA) and conversational instruction following (Sea-Bench).
\end{abstract}

\section{Introduction}\label{sec:intro}



Recent scaling effort in foundation large language models \citep{gpt3_brown2020language,palm_chowdhery2022palm,bloom_scao2022bloom,llama_touvron2023} with massive pre-training data has enabled them to learn a broad range of natural language tasks through few-shot in-context learning, where a few input-output exemplars are concatenated to the test input to prompt the model to predict the output and no gradient update of the model is performed. While most LLMs are pre-trained with multilingual corpora in addition to the gigantic English corpus, and have been shown to demonstrate impressive abilities in other languages \citep{gpt3_brown2020language,palm_chowdhery2022palm,bloom_scao2022bloom}, they only excel in high-resource languages, such as French. Further, they may still require pivoting the inputs into English, that is, performing tasks in English first before reverting the response back to native outputs \citep{llm_are_mling_cot_shi2022language,xlt_not_all_created_equal_huang2023not}. Improving LLMs' abilities in extremely low-resource languages can be even more challenging, particularly where the data coverage is less than 0.0001\% \citep{bloom_scao2022bloom} or none at all \citep{llama_touvron2023}. 
We also found that the models may confusedly respond in a wrong language or struggle with low-resource non-latin scripts due to overly fragmented tokenization, where words are broken into many byte-level tokens.

\begin{figure*}[ht]
  \centering
  \subfloat[X$\rightarrow$En]{
  \label{fig:ldp_example:x_en}
  \begin{tikzpicture}[node distance=0.5cm, auto,]
    \node[] (cmd) {\bf $\mathcal{F}_{\rightarrow en}$};
    \small
    \node[below=0.55cm of cmd.west, anchor=west] (src2) {\boxlightblue{Chinese: \begin{CJK*}{UTF8}{gbsn}早上好\end{CJK*}}};
    \node[below=0.55cm of src2.west, anchor=west] (tgt2) {\boxlightblue{English: Good morning}};
    \node[below=0.55cm of tgt2.west, anchor=west] (src4) {\boxlightblue{French: Je suis désolé}};
    \node[below=0.55cm of src4.west, anchor=west] (tgt4) {\boxlightblue{English: I'm sorry}};
    \node[below=0.55cm of tgt4.west, anchor=west] (src5) {\boxlime{Igbo: {\fontencoding{T5}\selectfont Ịmụ igwe}}};
    \node[below=0.55cm of src5.west, anchor=west] (tgt5) {\boxlime{English:} \red{Machine learning}};
    \normalsize
    \node[below=0.6cm of tgt5.west, anchor=west] (eval) {\checkmark language, \checkmark translation};
    \draw[rounded corners] (-0.6, -4.3) rectangle (3.6, 0.5) {};
  \end{tikzpicture}
  }
  \subfloat[En$\rightarrow$X]{
  \label{fig:ldp_example:en_x}
  \begin{tikzpicture}[node distance=0.5cm, auto,]
    \node[] (cmd) {\bf $\mathcal{F}_{\rightarrow ig}$};
    \small
    \node[below=0.55cm of cmd.west, anchor=west] (src2) {\boxlightblue{English: Good morning}};
    \node[below=0.55cm of src2.west, anchor=west] (tgt2) {\boxlightblue{Chinese: \begin{CJK*}{UTF8}{gbsn}早上好\end{CJK*}}};
    \node[below=0.55cm of tgt2.west, anchor=west] (src4) {\boxlightblue{English: I'm sorry}};
    \node[below=0.55cm of src4.west, anchor=west] (tgt4) {\boxlightblue{French: Je suis désolé}};
    \node[below=0.55cm of tgt4.west, anchor=west] (src5) {\boxlime{English: Machine learning}};
    \node[below=0.55cm of src5.west, anchor=west] (tgt5) {\boxlime{Igbo:} \red{kuosha mashine}};
    \normalsize
    \node[below=0.6cm of tgt5.west, anchor=west] (eval) {\red{\xmark} language, \red{\xmark} translation};
    \draw[rounded corners] (-0.6, -4.3) rectangle (3.55, 0.5) {};
  \end{tikzpicture}
  }
  \subfloat[En$\rightarrow$X with back-translation]{
  \label{fig:ldp_example:en_x_BT}
  \begin{tikzpicture}[node distance=0.5cm, auto,]
    \node[] (cmd) {\bf $\mathcal{F}^{bt}_{\rightarrow ig}$};
    \small
    \node[below=0.55cm of cmd.west, anchor=west] (src3) {\boxlightblue{English: \purple{20 years ago}}};
    \node[below=0.55cm of src3.west, anchor=west] (tgt3) {\boxlightblue{Igbo: \blue{{\fontencoding{T5}\selectfont Afọ 20 gara aga}}}}
      (tgt3.east) edge[pil, bend right=45] node[right=0.4em] {\bf $\mathcal{F}_{\rightarrow en}$} (src3.east);
    \node[below=0.55cm of tgt3.west, anchor=west] (src4) {\boxlightblue{English: \purple{Good evening}}};
    \node[below=0.55cm of src4.west, anchor=west] (tgt4) {\boxlightblue{Igbo: \blue{{\fontencoding{T5}\selectfont Mgbede ọma}}}}
      (tgt4.east) edge[pil, bend right=45] node[right=0.4em] {\bf $\mathcal{F}_{\rightarrow en}$} (src4.east);
    \node[below=0.55cm of tgt4.west, anchor=west] (src5) {\boxlime{English: Machine learning}};
    \node[below=0.55cm of src5.west, anchor=west] (tgt5) {\boxlime{Igbo:} \red{{\fontencoding{T5}\selectfont Ịmụ igwe}}};
    \normalsize
    \node[below=0.6cm of tgt5.west, anchor=west] (eval) {\checkmark language, \checkmark translation};
    \draw[rounded corners] (-0.6, -4.3) rectangle (4.6, 0.5) {};
  \end{tikzpicture}
  }
  \caption{LDP prompting for unsupervised translation. (\ref{fig:ldp_example:x_en}) $\mathcal{F}_{\rightarrow en}$ translates from any language into English by concatenating the fixed \colorbox{lightblue}{linguistically-diverse shots} and \colorbox{lime}{input text} to prompt LLMs to generate the correct \red{translation}. (\ref{fig:ldp_example:en_x}) Similarly $\mathcal{F}_{\rightarrow ig}$ translates English into Igbo, but with low accuracy. (\ref{fig:ldp_example:en_x_BT}) $\mathcal{F}^{bt}_{\rightarrow ig}$ translates English to Igbo using \colorbox{lightblue}{synthetic intra-lingual exemplars} generated from \blue{unlabeled target-language data} with $\mathcal{F}_{\rightarrow en}$.} 
  \label{fig:ldp_example}  
\end{figure*}


In this work, we propose Linguistically-Diverse Prompting (LDP), a technique that promotes an LLM to perform generative tasks in low-resource languages by demonstrating few-shot exemplars in a diverse set of high-resource languages. This method works in both unsupervised setup with foundation base LLMs \citep{bloom_scao2022bloom,llama_touvron2023} and pseudo-zero-shot setup instruction-tuned counterparts \citep{instructgpt_ouyang2022training,bloomz_muennighoff2022crosslingual,gpt4}, by synthetically creating few-shot examples from zero-shot prompting. An example of LDP for unsupervised translation task is shown in \Cref{fig:ldp_example}, where we gather a small set of synthetic $X$$\rightarrow$En exemplars from a diverse set of high-resource languages using a pretrained unsupervised MT model \citep{criss2020}. Then, we concatenate them as input-output few-shot prompts to illicit the LLM to produce translation in low-resource languages. Meanwhile, \Cref{sec:method}, along with \Cref{fig:ldp_directions}, explains LDP in other generalized adoptions in many other tasks. Our method is based on the following empirical observations of LLMs: \Ni in-context exemplars may play a larger role in helping the model \textit{locate} the task in its pre-trained knowledge \citep{icl_bayesian_infer_xie2021explanation}, \Nii LLMs possess dominant abilities in English while they may lag behind in other lower-resource languages \citep{instructgpt_ouyang2022training,llama_touvron2023,xlt_not_all_created_equal_huang2023not,llm_are_mling_cot_shi2022language}.





Our method is shown to perform on par with supervised prompting in unsupervised translation tasks between English and 13 Indic and 21 African low-resource languages, with BLOOM \citep{bloom_scao2022bloom} and InstructGPT (text-davinci-003) \citep{instructgpt_ouyang2022training} models. Furthermore, adapting our method to X$\rightarrow$Y non-English directions even outperforms supervised promptings by up to 3 chrF++ in pairs involving low-resource languages. In multilingual summarization tasks \citep{xsum_Narayan2018DontGM}, our zero-shot LDP method outperforms both basic prompting and other English-pivoting methods by up to 4 ROUGE-L and is generally favored by GPT-4-EVAL \citep{gpt_eval_liu2023g-eval}. With GPT-3.5, our method considerably improve performance of zero-shot question answering XQUAD \citep{xquad_2019} and TydiQA \citep{tqdiqa_clark2020tydi} tasks in 7 languages. Our method can even enable Llama-2 \textbf{base} \citep{llama2_touvron2023llama} to perform conversational instruction following tasks and improve the chat model in Sea-Bench \citep{seallm_nguyen2023seallms} for 2 languages that were not instruction-tuned.

\section{Related Work}\label{sec:rel_works}

\makeatletter
\pgfplotsset{
  col chart axis style/.style={
    width=\columnwidth,
    height=0.15\textheight,
    xtick=data,
    enlarge y limits=false,
    enlarge x limits=0.05,
  },
  col bar axis style/.style={
    col chart axis style,
    enlarge y limits=false,
    enlarge x limits=0.05,
    ymin=40, 
    ymax=65,
    ybar=0.1em,
    bar width=9pt,
    legend style={
      font=\footnotesize,
      legend columns=2,
      at={(0.15, 0.95)},  
      anchor=north, 
      /tikz/every even column/.append style={column sep=0.5cm},
    },
  },
}


Large language models (LLMs) display outstanding capabilities because they are pre-trained on massive amounts of internet text data \citep{gpt3_brown2020language,palm_chowdhery2022palm,bloom_scao2022bloom,llama_touvron2023,llama2_touvron2023llama}. Without any gradient update, base LLMs are able to perform in-context learning by simply observing a list of high-quality input-output exemplars \citep{gpt3_brown2020language,llm_do_icl_differently_wei2023larger}. This technique works across many tasks that involve language understanding, reasoning and generation \citep{gpt3_brown2020language,cot_wei2022chain,llm_are_mling_cot_shi2022language}. Much research has been conducted to understand in-context learning. 
Some suggest that the models secretly perform gradient descent on the exemplars \citep{icl_gradient_descen_dai2022can}, while others demonstrate that most of the knowledge is learned during pre-training, and the exemplars are only to provide evidence for the model to \textit{locate} the intended task via a Bayesian inference process \citep{icl_bayesian_infer_xie2021explanation,rethink_demonstration_min2022rethinking,lima_zhou2023lima}.



Most LLMs are trained with multilingual corpora \citep{ccnet_wenzek-etal-2020-ccnet}, even if these make up a tiny fraction of the large English corpora \citep{gpt2_radford2019language_gpt2,gpt3_brown2020language}. Despite that, LLMs still exhibit strong multilingual capabilities with high-resource languages like French, German and Chinese, often with the help of English-pivoting using supervised translation systems \citep{llm_are_mling_cot_shi2022language,few_shot_multilingual_llm} or prompting the model to firstly generate intermediate English context \citep{xlt_not_all_created_equal_huang2023not}. 
BLOOM \citep{bloom_scao2022bloom} is one of the LLMs trained with the most number of languages, whose ROOTS corpus consists of data from 46 languages \citep{roots_corpus}. 
The ROOTS corpus includes 34 Indic and African languages regarded as low-resource, with each language having a pre-training coverage of less than 1\% in Hindi for the Indic group, to $2e^{-5}$\% in Tumbuka for the African group, as shown in \Cref{fig:bloom_lang_coverage} in the Appendix. More recent multilingual models, like SeaLLM or Aya, were also introduced with better instruction-following abilities \citep{seallm_nguyen2023seallms,aya_model}. Ad-hoc multilingual pre-training approaches, like word-pair mining, have been explored \citep{improve_low_res_mllm}. Meanwhile, other works inspect the evaluation suites \citep{on_multilingual_capabilities} and the tokenization process for multilingual LLMs \citep{tokenization_impact_multilingual_lm}.
Exclusively only for unsupervised translation tasks, our linguistically-diverse prompting (LDP) strategy is also an English-pivoting method, but it is different from other cross-lingual counterparts \citep{llm_are_mling_cot_shi2022language,xlt_not_all_created_equal_huang2023not} in that while others only pivot inputs to English intermediates, we use in-context pairs between English and a diverse set of high-resource languages to promote the intended task in the target low-resource language. For other intra-lingual tasks like instruction following, where input and output are both expected to be in the same language, our LDP helps prevent English-tuned models \citet{llama2_touvron2023llama} from responding with English answer given a non-English query.



Part of our work also intersects with unsupervised multilingual machine translation (UMT), where back-translation is proven to be effective \citep{understanding_backtranslation_scale,lample2018phrase_unsup,conneau2019cross_xlm,mbart_liu2020multilingual,refine_lowres_umt_nguyen2022}, along with other methods \citep{criss2020,swavumt_nguyen2022contrastive}. English-pivoting is also prominent in the realm of machine translation, where training models on high-resource En$\leftrightarrow$X bitext improves lower-resource En$\leftrightarrow$Y tasks \citep{multilingual_view_garcia-etal-2020,harness-multilingual-garcia-etal-2021}. 
Analyses of machine translation using LLMs have also been done. \citet{howgood_gpt_mt_hendy2023good} show that GPT models can perform competitively alongside the best MT models. \citet{mmt_llm_analysis_zhu2023multilingual} focus on optimizing supervised exemplars selection strategies, while \citet{icl_maintain_coherence_sia2023context} discover that using specific coherent prompts for each input helps improve performance. Nonetheless, such studies only consider supervised instruction-tuned models \citep{instructgpt_ouyang2022training,bloomz_muennighoff2022crosslingual}, which may risk test-set contamination \citep{bloomz_muennighoff2022crosslingual}. Thus, there is still limited research involving low-resource languages in completely zero-shot setups. 
As such, since low-resource languages may not enjoy the privilege of having large unlabeled data to conduct searching, only random selection is used in this study, while optimal exemplar selection is out of scope.

\section{Method}\label{sec:method}





\subsection{Linguistically-Diverse Prompting (LDP)}\label{subsec:ldp}

Our method is inspired from two empirical observations: \Ni LLMs may have already learned most of the task concepts implicitly during pre-training, and that in-context exemplars play a larger role in providing evidence for the model to identify the intended task \citep{icl_bayesian_infer_xie2021explanation,rethink_demonstration_min2022rethinking,lima_zhou2023lima}. \Nii LLMs perform generative tasks dominantly well in only a handful of major languages (English and other high-resource ones), whose pre-training data is significantly abundant \citep{gpt3_brown2020language,llama_touvron2023,gpt4}. To achieve better performance on lower-resource languages, it has been shown that we may need to instruct the LLMs to generate intermediate reasoning in English before producing the final answers in the target language \citep{xlt_not_all_created_equal_huang2023not}; or to translate non-English inputs perform tasks in English entirely \citep{llm_are_mling_cot_shi2022language}.

\Cref{fig:ldp_example} illustrates how our LDP method aims to take advantage of the aforementioned observations in the case of unsupervised translation tasks.
Particularly, we prompt the model to identify the task of ``translating from \textit{any language $X$} into $E$'', by demonstrating pairs from ``every language'' to $E$. Practically, shown in \Cref{fig:ldp_example:x_en}, we use synthetic pairs from diverse high-resource languages as exemplars to prompt the models to translate the target low-resource language $X$ (\eg\ Igbo) into English (En) with high quality. Such diverse set of prompt languages should include various script types ranging from Latin alphabets to logograms. 
\Cref{fig:ldp_example:en_x} shows that applying the same technique for En$\rightarrow$$X$ task may results in incorrect translation.
In \Cref{fig:ldp_example:en_x_BT}, however, we leverage LDP to translate unlabeled texts of target $X$ language into En, forming back-translated synthetic pairs to prompt the model to translate from En to $X$ with higher quality. This is because the target-side distribution is now realistic and consistently close to the true target distribution, which has been shown to be crucial for in-context learning \citep{icl_bayesian_infer_xie2021explanation}.


\begin{figure*}[t]
  \centering
  \begin{subfigure}[t]{0.62\textwidth}
  \centering
  \begin{tikzpicture}[node distance=0.5cm, auto,]
    \node[] (cmd) {\bf $\mathcal{F}^{mt}_{\text{x}\rightarrow \text{en}}$};
    \texttt{
    \node[below=0.60cm of  cmd.west, anchor=west] (src1) {\boxpink{[fr]}\boxyellow{[en]}};
    \node[below=0.55cm of src1.west, anchor=west] (tgt1) {\boxpink{[vi]}\boxyellow{[en]}};
    \node[below=0.55cm of tgt1.west, anchor=west] (src2) {\boxpink{[zh]}\boxyellow{[en]}};
    \node[below=0.55cm of src2.west, anchor=west] (tgt2) {\boxlime{[x]} \red{{[$\hat{\text{en}}$]}}};
    }
    \draw[rounded corners] (-0.6, -2.7) rectangle (1.6, 0.5) {};
  \end{tikzpicture}
  \begin{tikzpicture}[node distance=0.5cm, auto,]
    \node[] (cmd) {\bf $\mathcal{F}^{mtbt}_{\text{en}\rightarrow\text{x}}$};
    \texttt{
    \node[below=0.60cm of  cmd.west, anchor=west] (src1) {\boxyellow{[en$_1^{\text{x}_1}$]}\boxpink{[x$_1$]}};
    \node[below=0.55cm of src1.west, anchor=west] (tgt1) {\boxyellow{[en$_2^{\text{x}_2}$]}\boxpink{[x$_2$]}};
    \node[below=0.55cm of tgt1.west, anchor=west] (src2) {\boxyellow{[en$_3^{\text{x}_3}$]}\boxpink{[x$_3$]}};
    \node[below=0.55cm of src2.west, anchor=west] (tgt2) {\boxlime{[en]} \red{{[$\overline{\text{x}}$]}}};
    }
    \draw[rounded corners] (-0.6, -2.7) rectangle (1.8, 0.5) {};
  \end{tikzpicture}
  \begin{tikzpicture}[node distance=0.5cm, auto,]
    \node[] (cmd) {\bf $\mathcal{F}^{mt}_{\text{x}\rightarrow \text{y}}$};
    \texttt{
      \node[below=0.60cm of  cmd.west, anchor=west] (src1) {\boxpink{[x$_1$]}\boxyellow{[$\hat{\text{en}}_1^{\text{x}_1}$]}\boxlightyellow{[$\overline{\text{y}}_1^{\text{en}_1}$]}};
      \node[below=0.55cm of src1.west, anchor=west] (tgt1) {\boxpink{[x$_2$]}\boxyellow{[$\hat{\text{en}}_2^{\text{x}_2}$]}\boxlightyellow{[$\overline{\text{y}}_2^{\text{en}_2}$]}};
      \node[below=0.55cm of tgt1.west, anchor=west] (src2) {\boxpink{[x$_3$]}\boxyellow{[$\hat{\text{en}}_3^{\text{x}_3}$]}\boxlightyellow{[$\overline{\text{y}}_3^{\text{en}_3}$]}};
      \node[below=0.55cm of src2.west, anchor=west] (tgt2) {\boxlime{[x]} \red{[en][y]}};
      }
      \draw[rounded corners] (-0.5, -2.7) rectangle (3.1, 0.5) {};
    \end{tikzpicture}
  \caption{LDP for translation for $X$$\rightarrow$En, En$\rightarrow$$X$ and $X$$\rightarrow$$Y$.}
  \label{fig:ldp_mt}
  \end{subfigure}
  \hfill
  \begin{subfigure}[t]{0.37\textwidth}
    \centering
  \begin{tikzpicture}[node distance=0.5cm, auto,]
    \node[] (cmd) {\bf $\mathcal{F}^{in}_{\text{x}}$ };
    \texttt{
    \node[below=0.60cm of  cmd.west, anchor=west] (src1) {\colorbox{pink}{[q$_\text{fr}$]}\colorbox{yellow}{[r$_\text{fr}$]}};
    \node[below=0.55cm of src1.west, anchor=west] (tgt1) {\colorbox{pink}{[q$_\text{vi}$]}\colorbox{yellow}{[r$_\text{vi}$]}};
    \node[below=0.55cm of tgt1.west, anchor=west] (src2) {\colorbox{pink}{[q$_\text{zh}$]}\colorbox{yellow}{[r$_\text{zh}$]}};
    \node[below=0.55cm of src2.west, anchor=west] (tgt2) {\colorbox{lime}{[q$_\text{x}$]} \red{[$\hat{\text{r}}_\text{x}$]}};
    }
    \draw[rounded corners] (-0.6, -2.7) rectangle (1.8, 0.5) {};
  \end{tikzpicture}
  \begin{tikzpicture}[node distance=0.5cm, auto,]
    \node[] (cmd) {\bf $\hat{\mathcal{F}}^{in}_{\text{x}}$ };
    \texttt{
    \node[below=0.60cm of  cmd.west, anchor=west] (src1) {\colorbox{pink}{[q$_\text{x}^1$]}\colorbox{yellow}{[$\hat{\text{r}}_\text{x}^1$]}};
    \node[below=0.55cm of src1.west, anchor=west] (tgt1) {\colorbox{pink}{[q$_\text{x}^2$]}\colorbox{yellow}{[$\hat{\text{r}}_\text{x}^2$]}};
    \node[below=0.55cm of tgt1.west, anchor=west] (src2) {\colorbox{pink}{[q$_\text{x}^3$]}\colorbox{yellow}{[$\hat{\text{r}}_\text{x}^3$]}};
    \node[below=0.55cm of src2.west, anchor=west] (tgt2) {\colorbox{lime}{[q$_\text{x}$]} \red{[r$_\text{x}$]}};
    }
    \draw[rounded corners] (-0.6, -2.7) rectangle (1.7, 0.5) {};
  \end{tikzpicture}
  \caption{LDP for intra-lingual tasks.}
  \label{fig:ldp_sum}
  \end{subfigure}
  \caption{Illustrations LDP for $X$$\rightarrow$En, En$\rightarrow$$X$ and $X$$\rightarrow$$Y$ cross-lingual translation (\ref{fig:ldp_mt}) and general intra-lingual tasks (\ref{fig:ldp_sum}). For $X$$\rightarrow$En, the colored box \boxpink{[z]} represents an unlabeled text in language z, \boxyellow{[en]} represents its corresponding En translation, while \boxlime{[x]} stands for the test input in language x and uncolored box \red{{[$\hat{\text{en}}$]}} represents model outputs. For En$\rightarrow$$X$, \boxyellow{[en$^{\text{x}}$]} represents En text translated with $\mathcal{F}^{mt}_{\text{x}\rightarrow \text{en}}$. For $X$$\rightarrow$$Y$, \boxlightyellow{[$\overline{\text{y}}^{\text{en}}$]} represents a text in language y translated from En text \boxyellow{[$\hat{\text{en}}^{\text{x}}$]}. Similarly for intra-lingual tasks like summarization (\ref{fig:ldp_sum}), \colorbox{yellow}{[$\hat{\text{r}}_\text{z}$]} represents a response in language z for query \colorbox{pink}{[q$_\text{z}$]}. 
  }
  \label{fig:ldp_directions}
\end{figure*}


\subsection{LDP for Cross-lingual Tasks (Translation)}\label{subsec:ldp_for_mt}

For tasks where the input and output are in different languages, such as translation, we adopt LDP for $X\rightarrow E$, $E\rightarrow X$ and $X\rightarrow Y$ (where $X, Y \neq E$), differently, as shown in \Cref{fig:ldp_mt}, where we assume $E=\text{English (En)}$ for better understanding.

\paragraph{$X\rightarrow E$ task.}As mentioned above, we first gather $n$ $Z_i$$\rightarrow$$E$ exemplar pairs $(z_i,e^{z_i})$ with $Z_i \in \mathcal{Z}$ and $\mathcal{Z}$ being a diverse set of languages with various writing systems, lexical and regional characteristics, such as Chinese (Zh), and $Z_i \notin \{X,E\}$. Such exemplars can be collected from unlabeled data $z_i$ of the respective language $Z_i$ and using unsupervised MT models to translate them into $E$ (for unsupervised tasks) as $e^{z_i}$, or from labeled few-shot pairs (for zero-shot tasks). From that, we can perform translation of an input $x$ of language $X$ into $E$ with an LLM ($\theta$) by conditioning the LDP prompts as:
\begin{equation}
  \mathcal{F}^{mt}_{X\rightarrow E}(x) \sim p_{\theta}(\cdot|x,z_1,e^{z_1},..,z_n,e^{z_n})\label{eqn:ldp_mt_x2en}
\end{equation}

\paragraph{$E\rightarrow X$ task.}We leverage $\mathcal{F}^{mt}_{X\rightarrow E}$ to build intra-lingual prompts with unlabeled data from the target language $X$. Specifically, given $m$ unlabeled texts $x_j \in \mathcal{D}_X$ with $\mathcal{D}_X$ being a monolingual corpus in language $X$, we produce synthetic back-translation (BT) target $e^X_j = \mathcal{F}^{mt}_{X\rightarrow E}(x_j)$. Then, we use the BT synthetic pairs as exemplars for $E\rightarrow X$ tasks for a test input $e$:
\begin{equation}
  \mathcal{F}^{mtbt}_{E\rightarrow X}(e) \sim p_{\theta}(\cdot|e,e^X_1,x_1,...,e^X_m,x_m)\label{eqn:ldp_mt_en2x}
\end{equation}
The intra-lingual exemplars with the same language in the target side helps the model locate the intended language to generate more effectively than a standard language tag, as these exemplars show the model \textit{what the intended language looks like}. 

Note that we could also use $\mathcal{F}^{mtbt}$ for $X\rightarrow E$ ($\mathcal{F}^{mtbt}_{X\rightarrow E}$) by simply swapping the direction of the $(e^X_j,x_j)$ to $(x_j,e^X_j)$.
However, we found in the experiments that both $\mathcal{F}^{mt}$ and $\mathcal{F}^{mtbt}$ perform similarly and on par with supervised prompting for the $X\rightarrow E$ task, suggesting that we do not need any supervised or unlabeled data to translate any language into English. Furthermore, in \Cref{subsec:ablation_study}, we demonstrate that we can even omit these back-translation exemplars entirely with non-BT $\mathcal{F}^{mt}$ LDP by using native language tags.

\paragraph{$X\rightarrow Y$ task.}We leverage $\mathcal{F}^{mtbt}_{X\rightarrow E}$ and $\mathcal{F}^{mtbt}_{E\rightarrow X}$ to build $E$-pivoting triplets from unlabeled text from the source side.
Specifically, given unlabeled text $x_j \in \mathcal{D}_X$ in language $X$, we back-translate them into $e^X_j=\mathcal{F}^{mtbt}_{X\rightarrow E}(x_j)$ of language $E$, which we then use to produce $y^E_j=\mathcal{F}^{mtbt}_{E\rightarrow Y}(e^X_j)$ of language $Y$.
This process forms triplets $[x_j,e^X_j,y^E_j]$ to prompt the model to generate intermediate $E$ translation before producing the final result in $Y$. Formally, given an input $x$, the translation in $Y$ is computed as:
\begin{equation}
  \mathcal{F}^{mt}_{X\rightarrow Y}(x) \sim p_{\theta}(\cdot|x,x_1,e^X_1,y^E_1,...,x_n,e^X_n,y^E_n)\label{eqn:ldp_mt_x2y}
\end{equation}



\paragraph{Unsupervised fine-tuning.}The $\mathcal{F}^{mt}_{X\rightarrow E}$ prompting method also allows us to create larger-scale synthetic $X$-$E$ data from unlabeled corpora to fine-tune the model for translation tasks without any in-context prompt at inference time. Specifically, we use the \texttt{[input]<lang-tag>[output]} template to construct multilingual training samples with the generated data pairs from multiple low-resource languages. We fine-tune the query-key-value linear weights of all attention layers, which account for 20-30\% of the total parameters to avoid overfitting.

\subsection{LDP for intra-lingual tasks}\label{subsec:ldp_for_sum}

For intra-lingual tasks, where the input and output are expected to be in the same language, such as summarization, question answering and instruction following, we adopt LDP in zero-shot setups for instruction-tuned models \citep{instructgpt_ouyang2022training} differently as illustrated in \Cref{fig:ldp_sum}. Formally, given a query $q_X$ in the target language $X$ and $n$ in-domain queries $q_{Z_i}$ with $Z_i \in \mathcal{Z}$ and $\mathcal{Z}$ being a diverse set of high-resource languages, we use standard or augmented zero-shot prompting strategies $h$ \citep{xlt_not_all_created_equal_huang2023not,cot_wei2022chain} to obtain responses $r_{Z_i} = h(q_{Z_i})$. We then use the synthetic query-response pairs $(q_{Z_i},r_{Z_i})$ as LDP in-context exemplars to compute the target-language response $r_X$ for $q_X$ as:
\begin{equation}
  \mathcal{F}_X^{in}(q_X) \sim p_{\theta}(y|q_X,q_{Z_1},r_{Z_1},..,q_{Z_n},r_{Z_n}) \label{eqn:ldp_sum}
\end{equation}
Similar to $E\rightarrow X$ translation task, we then use zero-shot $\mathcal{F}_X^{in}$ to generate synthetic intra-lingual prompts from $m$ unlabeled queries $q^j_X \in \mathcal{D}_X$ by producing responses $r^j_X = \mathcal{F}_X^{in}(q^j_X)$ in $X$ language. After that, we compute the final response for the input $q_X$ with $\hat{\mathcal{F}}_X^{in}$ as:
\begin{equation}
  \hat{\mathcal{F}}_X^{in}(q_X) \sim p_{\theta}(y|q_X,q^1_{X},r^1_{X},..,q^m_{X},r^m_{X}) \label{eqn:ldp_sum_bt}
\end{equation}


\section{Experiments}\label{sec:experiments}

In this section, we evaluate our method in various translation (\Cref{subsec:lowres_mt,subsec:x2y_mt}),  summarization (\ref{subsec:summarization}), question answering (\ref{subsec:qa}) and instruction-following (\ref{subsec:instruction_following}) across different settings and languages. 
We also conduct extensive analyses to provide further insights into our method (\ref{subsec:ablation_study}).

\begin{table}[t]
  \centering
  \resizebox{\columnwidth}{!}{%
  \setlength{\tabcolsep}{1.4pt}
  \begin{tabular}{lcccc}
  \hline
    & {\bf Ind-En}  & {\bf En-Ind}  & {\bf Afr-En}  & {\bf En-Afr} \\
  \hline
  \multicolumn{5}{l}{\bf Base BLOOM-175B	}\\
  Supervised-8-shot                    & 47.31 & 34.66 & 28.64 & 14.93 \\
  Unsupervised-LDP                     & 47.62 & 34.54 & 28.72 & 14.57 \\
  \hline
  \multicolumn{5}{l}{\bf Base BLOOM-7B	}\\
  Supervised-8-shot                    & 39.86 & 24.02 & 21.51 & 11.27 \\
  Unsupervised-LDP                     & 39.88 & 24.41 & 20.47 & 12.04 \\
  Fine-tune                            & 42.19 & 32.72 & 21.14 & 15.73 \\ 
  \hline
  \multicolumn{5}{l}{\bf Supervised InstructGPT (text-davinci-003)} \\
  Zero-shot                            & 35.37 & 20.71 & 27.10 & 15.45  \\
  Supervised-6-shot                    & 37.07 & 24.74 & 31.51 & 19.22  \\
  Unsupervised-LDP                     & 38.45 & 25.17 & 31.92 & 19.51  \\
  \hline
  \multicolumn{5}{l}{\bf Supervised upperbound}\\
  NLLB-200 distilled                   &  \textit{61.00} & \textit{46.77} & \textit{48.42} & \textit{39.18} \\
  \hline
  \end{tabular}
  }
  \caption{
    Averaged performances of different prompting techniques across various model sizes and types, namely BLOOM \citep{bloom_scao2022bloom} and InstructGPT text-davinci-003 \citep{gpt3_brown2020language,instructgpt_ouyang2022training}, in translation tasks between English (En) and 13 Indic (Ind) and 21 African (Afr) low-resource languages present in the ROOTS corpus \citep{roots_corpus}. SacreBLEU scores are provided in the Appendix.
  }
  \label{table:main_unsup_llmmt}
\end{table}

\subsection{Low-resource $\leftrightarrow$ English Translation}\label{subsec:lowres_mt}


As the ROOTS corpus \citep{roots_corpus} that BLOOM \citep{bloom_scao2022bloom} was pre-trained on offers the most diverse language coverage with open-sourced transparency, we tested our methods mainly with the BLOOM model on 13 Indic (Ind) languages and 21 African (Afr) languages present in the ROOTS corpus. We also conduct experiments with supervised InstructGPT (text-davinci-003) \citep{instructgpt_ouyang2022training} to provide further references. As not much detail about text-davinci-003 has been disclosed, its results are only to compare prompting techniques within the same model and not between models. Specific details about languages and test sets are provided in the {Appendix}. Following \citet{nllb_costa2022no_flores200}, we report results in mainly chrF++ \citep{popovic2015chrf}, which is a universal metric for all languages, while also reporting SacreBLEU \citep{sacredbleu_post-2018-call} in the Appendix.


In terms of methodologies, for supervised prompting, we collect as many supervised pairs as the models can fit within their context lengths (8 for BLOOM and 6 for GPT davinci-003). 
We use \texttt{<src>[input]\textbackslash n<tgt>[output]} as the prompt template, where \texttt{<src>} and \texttt{<tgt>} are the language tag names in English.
For our unsupervised linguistically-diverse prompting (LDP) method, we use 4 LDP $Z_i$$\leftrightarrow$En pairs from Arabic (Ar), Chinese (Zh), Vietnamese (Vi) and French (Fr) to conduct $X\rightarrow E$ synthetic data generation with $\mathcal{F}^{mt}_{X\rightarrow E}$ before using them as intra-lingual prompts for the target pair with $\mathcal{F}^{mtbt}_{X\leftrightarrow E}$, as explained in \Cref{sec:method}. For LDP, we do not include the language tags in the prompts as they offer no benefit. In our fine-tuning experiment, we use $\mathcal{F}^{mt}_{X\rightarrow E}$ to generate synthetic training data from various unlabeled sources \citep{ccnet_wenzek-etal-2020-ccnet} to fine-tune BLOOM-7B. 

\begin{table*}[t]
  \centering
  \resizebox{\textwidth}{!}{%
  \begin{tabular}{l|cc|cccc|cccc}
    \hline
    &  \multicolumn{2}{c}{\bf High-High}			&	\multicolumn{4}{c}{\bf High-Low}	 &  \multicolumn{4}{c}{\bf Low-Low}\\
                              & Vi-Fr	& Fr-Vi	& Zh-Ne	& Ne-Zh	& Es-Pa	& Pa-Es	& Ta-Sw	& Sw-Ta	& Te-Sw	& Sw-Te	\\
    \hline
    \multicolumn{11}{l}{\bf Foundation BLOOM-175B	}\\
    Supervised-8-shot	        & 52.17	&	51.50	&	30.91	&	17.83	&	25.67	&	37.71	&	31.45	&	31.81	&	31.46	&	25.84	\\
    Unsupervised-LDP        	& 52.66	&	50.24	&	31.61	&	18.34	&	27.85	&	39.51	&	34.61	&	34.47	&	32.14	&	30.57	\\
    \hline
    \multicolumn{11}{l}{\bf Supervised InstructGPT (text-davinci-003)}\\
    XLT \citep{xlt_not_all_created_equal_huang2023not}  & 51.16	& 44.84	& 28.56	&	13.26	&	23.61	&	34.18	&	24.20	& 25.46	&	24.89	&	23.48	\\
    Unsupervised-LDP                                    & 51.19	& 45.80	&	28.67	&	15.80	&	25.40	&	35.02	&	27.24	&	27.70	&	28.95	&	25.12	\\
    \hline
  \end{tabular}
  }
  \caption{
    chrF++ translation scores for X$\rightarrow$Y non-English tasks across high-high, high-low and low-low groups. 
  }
  \label{table:x2y}
\end{table*}

\Cref{table:main_unsup_llmmt} shows the averaged chrF++ scores for translations between English and 13 Indic and 21 African low-resource languages across different prompting techniques with various models. Noticeably, our unsupervised-LDP method performs on par with supervised prompting across all language groups and LLM models. This indicates that the synthetic prompts generated by our $\mathcal{F}^{mt}_{X\rightarrow E}$ technique are as good as supervised prompts when serving as few-shot exemplars,\footnote{The synthetic outputs themselves are still lower-quality than supervised translations or the ground truths.} thanks to the LLMs' outstanding ability in English. Furthermore, fine-tuning a 7B model with data generated by itself helps the model to advance towards the performance of its 175B sibling, especially for En$\rightarrow$$X$ direction. This suggests that fine-tuning the model on more low-resource language data improves generative abilities in such languages. 

For text-davinci-003, we observe the same pattern when comparing supervised and unsupervised-LDP. It is interesting to see that GPT's scores for Indic languages are lower than BLOOM but higher for African languages, despite the fact that the African languages are likely to have less data coverage. 
One of the reasons may be the token fragmentation issue which we explain in the Appendix. Similarly, we observe LDP performs competitively with supervised prompting on 20 European languages with LLaMA \citep{llama_touvron2023}, which we also detail in \Cref{table:unsup_llama} in the Appendix.


\subsection{Non-English-centric Translation}\label{subsec:x2y_mt}

For non-English $X$$\rightarrow$$Y$ directions, we compare our unsupervised method $\mathcal{F}^{mt}_{X\rightarrow Y}$ with supervised prompting in three categories: High-High resource languages with Vi and Fr, High-Low resource between Zh, Es, Ne (Nepali) and Pa (Punjabi), and Low-Low resource languages with Sw (Swahili), Ta (Tamil) and Te (Telugu). We use the same model and evaluation pipelines as explained \Cref{subsec:lowres_mt}. For this experiment, we evaluate on the FLoRes-200 devtest sets \citep{nllb_costa2022no_flores200}.
As reported in \Cref{table:x2y}, our unsupervised LDP technique also performs on par with supervised prompting in High-High Vi-Fr pairs. More interestingly, for High-Low and Low-Low language pairs, our unsupervised method even outperforms supervised prompting for these languages by up to 5 chrF++, largely thanks to the presence of English intermediate translations in the exemplars.

\subsection{Zero-shot Summarization}\label{subsec:summarization}


\begin{table}[h]
  \centering
  \begin{tabular}{l|c|c|c|c|c}
    \hline
            & {\bf Es} & {\bf Id} & {\bf Sw} & {\bf So} & {\bf Mr} \\
    \hline
    Basic	 & 2.9	& 2.5	& 2.3	& 3.0	& 2.9 \\
    XLT	     & 3.9	& 3.4	& 3.1	& 3.9	& 3.8 \\
    LDP	     & 4.1	& 3.6	& 3.3	& 4.0	& 3.9 \\
    LDP+U	 & \textbf{4.2} & \textbf{3.8} & \textbf{3.3}	& \textbf{4.0}	& \textbf{3.9} \\
    \hline
  \end{tabular}
  \caption{
    GPT-4-EVAL scores (1-5 ratings) of different prompting techniques using InstructGPT text-davinci-003 for zero-shot summarization in high-resource (Es, Id) and low-resource (Sw, So, Mr) in the XL-sum summarization task \citep{xsum_Narayan2018DontGM}. ROUGE-L scores are provided in the Appendix.
  }
  \label{table:xsum}
\end{table}

We extend our LDP method to multilingual summarization by combining intral-lingual LDP (\cref{subsec:ldp_for_sum}) with cross-lingual prompting (XLT) \citep{xlt_not_all_created_equal_huang2023not} using the supervised text-davinci-003 model. XLT is a recent English-pivoting instruction proposed by \citet{xlt_not_all_created_equal_huang2023not}. We follow the LDP adoptions for intral-lingual tasks with (LDP+U or $\hat{\mathcal{F}}^{in}$) and without (LDP or $\mathcal{F}^{in}$) unlabeled data, as described in \Cref{subsec:ldp_for_sum}. We conduct evaluation on the Extreme Summarization benchmark \citep{xsum_Narayan2018DontGM} in both high-resource (Es, Id-Indonesian) and low-resource (Sw, So-Somali, Mr-Marathi) languages. 
We evaluate the models with GPT-4-EVAL \citep{gpt_eval_liu2023g-eval} and ROUGE-L \citep{lin2004rouge}. GPT-4-EVAL is a GPT-4 based metric that recently scores best in human judgement alignment. We compare our methods with XLT, and basic instruction. 
As shown in \Cref{table:xsum}, our methods are consistently preferred by GPT-4-EVAL with higher ratings. In terms of ROUGE-L, whose scores are reported in the Appendix, our LDP methods also outperform standard XLT across all languages by up to 7 ROUGE-L and exceeds basic prompting by large margins.


\subsection{Zero-shot Question Answering}\label{subsec:qa}


\begin{table}[t]
  \centering
  \resizebox{\columnwidth}{!}{%
  \setlength{\tabcolsep}{2.5pt}
  \begin{tabular}{l|ccc|cccc}
    \hline
    \multirow{2}{*}{\bf GPT-3.5} &  \multicolumn{3}{c}{\bf XQUAD}			&	\multicolumn{4}{c}{\bf TydiQA}	\\
                & Ar	& Hi	& Th	& Ar	& Bn	& Fi	& Ru	\\
    \hline
    3-shot	                & 69.9	& 69.3	& 53.8	& 27.7	& 20.2	& 34.7	& 16.8 \\
    \hline
    0-shot	                & 52.9	& 45.9	& 26.3	& 19.1	&  5.7	& 21.7	& 12.3 \\
    \hspace{0.1em} w/ LDP   & 69.8	& 69.0	& 54.0	& 23.2	& 18.9	& 32.6	& 17.0 \\
  \end{tabular}
  }
  \caption{
    Multilingual question answering F1 scores of ChatGPT (GPT 3.5) using different prompting techniques across different languages in the XQUAD and TydiQA benchmarks.
  }
  \label{table:qa}
\end{table}

Our method also works well for multilingual comprehension and world-knowledge question answering with the XQUAD \citep{xquad_2019} and no-context TydiQA \citep{tqdiqa_clark2020tydi} benchmarks respectively. We demonstrate this with ChatGPT-3.5 across Arabic (Ar), Hindi (Hi), Thai (Th), Bengali (Bn), Finnish (Fi) and Russian (Ru). We select supervised exemplars from En, Vi, and Zh as LDP pairs for XQUAD and similarly exemplars from En, Id, Ko-Korean for TydiQA. As shown in \Cref{table:qa}, our method improves zero-shot and rivals 3-shot supervised prompting across various low-resource languages.

\subsection{General Instruction Following}\label{subsec:instruction_following}




\begin{table}[t]
  \centering
  \resizebox{\columnwidth}{!}{%
  \setlength{\tabcolsep}{2.5pt}
  \begin{tabular}{l|cc|cc|cc}
    \hline
    &  \multicolumn{2}{c}{\bf Task}				 &  \multicolumn{2}{c}{\bf Instruct} &  \multicolumn{2}{c}{\bf NatQA}  \\
                              & Vi & Id & Vi & Id & Vi & Id  \\
    \hline
    ChatGPT (3.5)	          & 7.47 & 7.85 & 9.42 & 9.80 & 9.05 & 9.45 \\
    \hline
    \multicolumn{5}{l}{ LLama2-13B}\\
    \hspace{0.1em}-Chat	        & 6.45 & 5.45 & 6.15 & 7.67 & 4.95 & 5.65 \\
    \hspace{0.1em}-Base w/ LDP	& 3.87 & 2.61 & 4.65 & 7.05 & 4.80 & 6.10 \\
    \hspace{0.1em}-Chat w/ LDP	& 3.83 & 6.54 & 8.57 & 8.72 & 4.94 & 6.85 \\
    \hline
  \end{tabular}
  }
  \caption{
    GPT-4 rated LLM-as-a-judge scores \citep{mt_bench_zheng2023judging} of different models and prompting strategies for task-solving (\textbf{Task}), instruction-following (\textbf{Instruct}) and natural question answering (\textbf{NatQA}) categories in the Sea-bench set \citep{seallm_nguyen2023seallms} for Vi and Id.
  }
  \label{table:instruct_following}
\end{table}

Beyond traditional NLP tasks, we also show that our LDP prompts can ellicit chatbot-style instruction following abilities in base pre-trained model \textbf{without} any supervised fine-tuning, and improve English-tuned models. Specifically, we utilize Sea-bench \citep{seallm_nguyen2023seallms} - a set of categorized instructions in multiple languages, designed to evaluate models with LLM-as-a-judge recipe \citep{mt_bench_zheng2023judging}. We measure GPT-4 rated scores of LLama2-13B base and chat models \citep{llama2_touvron2023llama}, using 4 random instructions from En, Zh, Fr, Ru as LDP prompts.
As shown in \Cref{table:instruct_following}, our method can invoke relatively good instruction following capability in Vi and Id even with a \textbf{base} model. With Llama2-chat, which has undergone supervised finetuning, our method can further improve the performance in various benchmarks for certain under-represented languages.



\makeatletter
\pgfplotsset{
  chart axis style/.style={
    width=\textwidth,
    height=0.16\textheight,
    xtick=data,
    enlarge y limits=false,
    enlarge x limits=0.05,
  },
  bar axis style/.style={
    chart axis style,
    ylabel=Accuracy (\%),
    xticklabel style={text height=1ex},
    enlarge y limits=false,
    enlarge x limits=0.05,
    ymin=40, 
    ymax=65,
    ybar=0.1em,
    bar width=9pt,
    legend style={
      font=\footnotesize,
      legend columns=2,
      at={(0.3, -0.2)},  
      anchor=north, 
      /tikz/every even column/.append style={column sep=0.5cm},
    },
  },
  line axis style/.style={
    chart axis style,
    axis lines*=right,
    xticklabels={},
    tickwidth=0pt,
    ymin=0,
    ymax=1.0,
    grid=none,
    legend style={
      font=\footnotesize,
      legend columns=2,
      at={(0.75, -0.2)},  
      anchor=north, 
      /tikz/every even column/.append style={column sep=0.5cm},
    },
  }
}
\makeatother


\begin{figure}[h]
  \centering
  \begin{subfigure}[b]{0.95\columnwidth}
      \includegraphics[width=\textwidth]{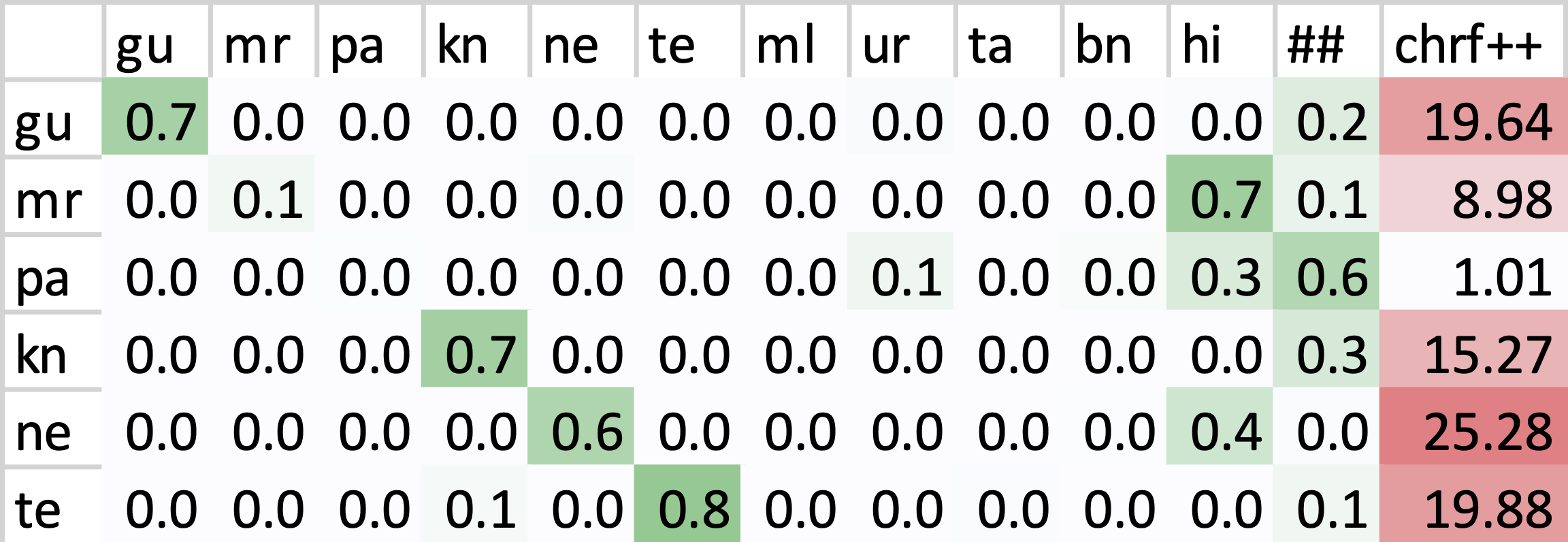}
      \caption{LDP without back-translation $\mathcal{F}^{mt}_{\text{En}\rightarrow X}$.}
      \label{fig:gen_right_language:without_bt}
    \end{subfigure}
    \hfill
    \begin{subfigure}[b]{0.95\columnwidth}
        \centering
      \includegraphics[width=\textwidth]{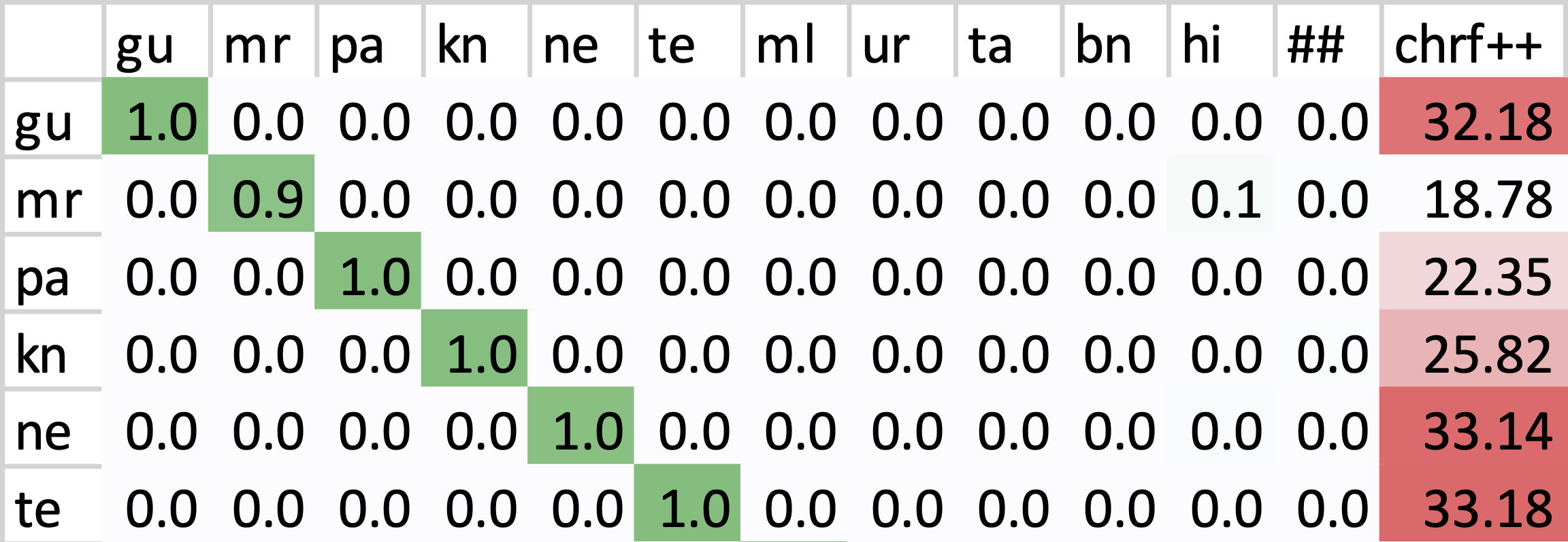}
      \caption{LDP with back-translation $\mathcal{F}^{mtbt}_{\text{En}\rightarrow X}$.}
      \label{fig:gen_right_language:with_bt}
  \end{subfigure}
  \caption{Probabilities of whether the BLOOM model generates the right language for En$\rightarrow$$X$ task using LDP without (\ref{fig:gen_right_language:without_bt}) and with (\ref{fig:gen_right_language:with_bt}) intra-lingual BT prompts. Columns indicate the languages the model generates into while rows are the languages it is \textit{supposed} to generate. \#\# are other languages.}
  \label{fig:gen_right_language}
\end{figure}

\subsection{Ablation Study}\label{subsec:ablation_study}

In this section, we conduct various analyses in the unsupervised translation tasks to provide a deeper understanding of our LDP method and the importance of each component, while presenting more experiments in the Appendix.


\paragraph{Generating the Right Language.}\Cref{fig:gen_right_language:without_bt} reveals one reason the models struggle to translate En$\rightarrow$$X$ when using LDP prompts $\mathcal{F}^{mt}$ (without intra-lingual BT data) is that the target-side distribution contains multiple languages, and the models struggle to recognize unfamiliar language tags, such as Marathi (Mr), and often generate wrong translations in the wrong languages (\eg\ Hindi instead of Marathi). Meanwhile, supplying synthetic intra-lingual prompts where the target-side is consistently in the intended language, as shown in \Cref{fig:gen_right_language:with_bt} with $\mathcal{F}^{mtbt}$, is more important in getting the models to recognize language rather than the language tag. In fact, we found that removing the language tag entirely can help improve the performance slightly.

\begin{table}[t]
  \centering
  \setlength{\tabcolsep}{2pt}
  \subfloat[Different language tags (chrF++).]{
  \label{table:ldp_language_tag}
  \begin{tabular}{l|cc}
    \hline
    {\bf BLOOM}  & {\bf Ind-En}  & {\bf En-Ind}  \\
    \hline
    \multicolumn{3}{l}{\bf Unsupervised LDP}\\
    En-tag               & 46.96	& 22.53	 \\
    En-tag + BT          & 47.43	& 34.41	 \\
    Native-tag           & 46.90	& 29.80	 \\
    Native-tag + BT      & 47.52	& 35.22	 \\
    No-tag               & 46.81	& --	   \\
    No-tag + BT          & 47.62  & 34.54 \\
    \hline
  \end{tabular}
  }
  \quad
  \subfloat[Choices of LDP languages (chrF++).]{
  \label{table:ldp_language_choice}
  \begin{tabular}{l|cccc}
  \hline
  {\bf BLOOM}  & {\bf Indic10-En}  & {\bf En-Indic10}  \\
  \hline
  {\bf Supervised}            & 46.32	& 32.44 \\
  \hline
  \multicolumn{3}{l}{\bf Unsupervised LDP with $\mathcal{Z}=$}\\
  Ar,Zh,Vi,Fr (default)       & 45.53	& 17.65 \\
  Hi,Hi,Hi,Hi (Hindi)         & 43.27	& 15.34 \\
  Ta,Bn,Hi (Indic)            & 45.51	& 16.25 \\
  Fr,Es,Pt (European)         & 45.31	& 18.98 \\
  Vi,Vi,Vi,Vi                 & 44.91	& 12.94 \\
  Zh,Zh,Zh,Zh                 & 44.71	& 15.78 \\
  Ar,Fr,Es,Pt,Vi,Zh,Id        & 45.50	& 16.88 \\
  \hline
  \end{tabular}
  }
  \caption{
    (\ref{table:ldp_language_tag}): Impact of English tag, native language tags and no language tag for in-context prompts in Indic languages. (\ref{table:ldp_language_choice}): Impact of different choices of LDP languages on $X$$\rightarrow$En directions using LDP without back-translation ($\mathcal{F}^{mt}$) across \textbf{10} Indic languages excluding Ta, Bn and Hi (Indic10). Note that we use supervised exemplars in \Cref{table:ldp_language_choice} for analysis purpose.
  }
  \label{table:which_ldp_languages_and_tags}
\end{table}


\paragraph{Impact of Native Language Tag.}The reason why we need unlabeled text to create intra-lingual prompts for En$\rightarrow$$X$ direction is because the models fail to recognize the correct language from the English language tags. A convenient way to eliminate such unlabeled text is to replace English-tag prompts (\eg\ ``\texttt{Spanish:[es-text]\textbackslash nChinese:[zh-text]}'') with native language tags for the target language (\eg\ ``\texttt{Española:[es-text]\textbackslash n}\begin{CJK*}{UTF8}{gbsn}中文\end{CJK*}:[zh-text]''). Such native tags serve as examples of how the intended language looks like.
As shown in \Cref{table:ldp_language_tag}, using LDP with native language tags without using any unlabeled text or intra-lingual back-translation (BT) prompts improves the performance of En$\rightarrow$$X$ tasks significantly, compared to using English tags. This method even approaches the performance of 8-shot supervised prompting and LDP with unlabeled BT prompts. Combining it with back-translation data (Native-tag + BT) even helps it outperform supervised prompting. In fact, the English tag may confuse the model to an extent that not using the language tag at all (\eg using ``\texttt{Input:[input]\textbackslash nOutput:[output]}'') does not hurt the performances.


\paragraph{Choice of LDP languages.}Another necessary question to ask is which high-resource languages should be selected as LDP exemplars. \Cref{table:ldp_language_choice} examines which LDP language choice is optimal. As shown, for 10 Indic low-resource languages, choosing a single related language (Hindi), which is often called cross-lingual prompting \citep{prompt_llm_mt_casestudy_zhang2023prompting,mmt_llm_analysis_zhu2023multilingual}, can be disastrous as the model tends to translate the prompt language rather than the test language. Choosing a single but distant language, like Vi, yields better results, while choosing a wide variety of languages across different regions (\eg\ Ar,Zh,Vi,Fr) may be the optimal choice.

\paragraph{Comparison with Unsupervised MT.}We also compare our method against the specialized unsupervised MT model CRISS \citep{criss2020} on eligible languages (Gu, Ne, Hi). As shown in \Cref{table:compare_criss}, unsupervised LDP prompting with BLOOM significantly outperforms CRISS across all languages, thanks to its larger size and strong English abilities.


\makeatletter
\pgfplotsset{
  split axis style/.style={
    width=\textwidth,
    height=0.2\textheight,
    xtick=data,
    enlarge y limits=false,
  },
  split line axis style/.style={
    ylabel=chrF++,
    split axis style,
    xticklabels={},
    tickwidth=0pt,
    ymin=0,
    ymax=1.0,
    grid=none,
    xtick=data,
    xticklabel style={text height=1ex,yshift=-1},
    yticklabel style={font=\small},
    legend style={
      font=\tiny,
      at={(0.02, 0.85)},  
      anchor=west, 
    },
  }
}
\makeatother

\begin{table}[h]
  \centering
  \resizebox{\columnwidth}{!}{%
  \setlength{\tabcolsep}{2.5pt}
  \begin{tabular}{l|cccccc}
    \hline
              & \multicolumn{2}{c}{\bf Gu-En} & \multicolumn{2}{c}{\bf Ne-En} & \multicolumn{2}{c}{\bf Hi-En} \\
              & $\rightarrow$ & $\leftarrow$ & $\rightarrow$ & $\leftarrow$ & $\rightarrow$ & $\leftarrow$ \\ 
    \hline
    CRISS                       & 41.88	 & 32.41 & 37.64 & 28.17 & 51.23 & 42.29	 \\
    \hline
    \multicolumn{5}{l}{\bf BLOOM Prompting}\\
    Supervised     & 51.63	 & 38.23 & 47.07 & 35.91 & 55.18 & 44.94 \\
    LDP            & 50.09	 & 37.63 & 48.26 & 35.76 & 55.71 & 45.36	 \\
    \hline
  \end{tabular}
  }
  \caption{Comparison in chrF++ between unsupervised LDP prompting and specialized unsupervised MT CRISS \citep{criss2020}}
  \label{table:compare_criss}
\end{table}

\begin{figure}[h]
\begin{tikzpicture}
  \begin{axis}[
    split line axis style,
    xticklabels from table={\datafinetuneindic}{params_n},
    height=0.17\textheight,
    width=\columnwidth,
    ymin=-1,
    ymax=10,
    legend style={
    font=\tiny,
    at={(0.0, 0.82)},  
    anchor=west, 
  },
  ]
    \addplot [blue, mark=*] table[x=id,y=x2en_diff] \datafinetuneindic;;
    \addplot [dashed, violet, mark=*] table[x=id,y=en2x_diff] \datafinetuneindic;;
    \draw [dashed, black, thick] (0,0) -- (8,0);
    \legend{Ind$\rightarrow$En, En$\rightarrow$Ind}
  \end{axis}
\end{tikzpicture}
\vspace{-0.5em}
\captionof{figure}{Gains achieved by fine-tuning BLOOM-7B w.r.t numbers of trainable parameters.}
\label{fig:finetune_vs_prompt}
\end{figure}

\paragraph{Fine-tuning Trainable Parameters.}\Cref{fig:finetune_vs_prompt} analyzes how LoRA-fine-tuned BLOOM-7B models \citep{lora_hu2021lora} perform in $X$$\rightarrow$En and En$\rightarrow$$X$ Indic translation tasks as we increase the trainable parameters from 30M to 2B (full query-key-value weights). As shown, gain margins for $X$$\rightarrow$En are relatively low within 1 chrF++ as we fine-tune more parameters. Meanwhile, we observe a substantial gain of 8.7 chrF++ for En$\rightarrow$$X$ task, suggesting that learning to generate an unfamiliar language needs much more parameters, rendering parameter-efficient methods, like LoRA, ineffective.

\section{Conclusion}

We introduce linguistically-diverse prompting (LDP), which is designed to use synthetic high-quality in-context exemplars from high-resource languages to prompt LLMs to perform generative tasks in low-resource languages. Our unsupervised approach achieves on par with supervised few-shot learning while using zero supervision in English to and from 34 low-resource Indic and African translation tasks, even outperforming supervised prompting in non-English-centric directions. Our method also outperforms other English-pivoting techniques in multilingual summarization.

\section{Limitations}

Our linguistically-diverse prompting method comes with a few limitations that should be considered when used. First, it is a way to invoke and improve LLM’s abilities in low-resource languages, and not necessarily boosting low-resource knowledge beyond the data the model was trained. Second, the presence of texts in the target low-resource languages are often needed in the context for the method to work effectively, thus it does not entirely eliminate the need for unlabeled data in such languages at inference times. Third, like many methods with LLMs, hallucinations may occur with our LDP prompting method. 

Regarding ethical impact, we do not foresee any potential ethical issues with our proposed method.

\bibliography{anthology,acl_latex}

\begin{thebibliography}{55}
\expandafter\ifx\csname natexlab\endcsname\relax\def\natexlab#1{#1}\fi

\bibitem[{Adelani et~al.(2022)Adelani, Alabi, Fan, Kreutzer, Shen, Reid, Ruiter, Klakow, Nabende, Chang, Gwadabe, Sackey, Dossou, Emezue, Leong, Beukman, Muhammad, Jarso, Yousuf, Niyongabo~Rubungo, Hacheme, Wairagala, Nasir, Ajibade, Ajayi, Gitau, Abbott, Ahmed, Ochieng, Aremu, Ogayo, Mukiibi, Ouoba~Kabore, Kalipe, Mbaye, Tapo, Memdjokam~Koagne, Munkoh-Buabeng, Wagner, Abdulmumin, Awokoya, Buzaaba, Sibanda, Bukula, and Manthalu}]{mafandmt_adelani-etal-2022-thousand}
David Adelani, Jesujoba Alabi, Angela Fan, Julia Kreutzer, Xiaoyu Shen, Machel Reid, Dana Ruiter, Dietrich Klakow, Peter Nabende, Ernie Chang, Tajuddeen Gwadabe, Freshia Sackey, Bonaventure F.~P. Dossou, Chris Emezue, Colin Leong, Michael Beukman, Shamsuddeen Muhammad, Guyo Jarso, Oreen Yousuf, Andre Niyongabo~Rubungo, Gilles Hacheme, Eric~Peter Wairagala, Muhammad~Umair Nasir, Benjamin Ajibade, Tunde Ajayi, Yvonne Gitau, Jade Abbott, Mohamed Ahmed, Millicent Ochieng, Anuoluwapo Aremu, Perez Ogayo, Jonathan Mukiibi, Fatoumata Ouoba~Kabore, Godson Kalipe, Derguene Mbaye, Allahsera~Auguste Tapo, Victoire Memdjokam~Koagne, Edwin Munkoh-Buabeng, Valencia Wagner, Idris Abdulmumin, Ayodele Awokoya, Happy Buzaaba, Blessing Sibanda, Andiswa Bukula, and Sam Manthalu. 2022.
\newblock \href {https://doi.org/10.18653/v1/2022.naacl-main.223} {A few thousand translations go a long way! leveraging pre-trained models for {A}frican news translation}.
\newblock In \emph{Proceedings of the 2022 Conference of the North American Chapter of the Association for Computational Linguistics: Human Language Technologies}, pages 3053--3070, Seattle, United States. Association for Computational Linguistics.

\bibitem[{Armengol-Estap{\'e} et~al.(2022)Armengol-Estap{\'e}, de~Gibert~Bonet, and Melero}]{on_multilingual_capabilities}
Jordi Armengol-Estap{\'e}, Ona de~Gibert~Bonet, and Maite Melero. 2022.
\newblock \href {https://aclanthology.org/2022.lrec-1.327} {On the multilingual capabilities of very large-scale {E}nglish language models}.
\newblock In \emph{Proceedings of the Thirteenth Language Resources and Evaluation Conference}, pages 3056--3068, Marseille, France. European Language Resources Association.

\bibitem[{Artetxe et~al.(2019)Artetxe, Ruder, and Yogatama}]{xquad_2019}
Mikel Artetxe, Sebastian Ruder, and Dani Yogatama. 2019.
\newblock \href {http://arxiv.org/abs/1910.11856} {On the cross-lingual transferability of monolingual representations}.
\newblock \emph{CoRR}, abs/1910.11856.

\bibitem[{Barrault et~al.(2020)Barrault, Biesialska, Bojar, Costa-juss{\`a}, Federmann, Graham, Grundkiewicz, Haddow, Huck, Joanis, Kocmi, Koehn, Lo, Ljube{\v{s}}i{\'c}, Monz, Morishita, Nagata, Nakazawa, Pal, Post, and Zampieri}]{wmt2020_barrault-etal-2020-findings}
Lo{\"\i}c Barrault, Magdalena Biesialska, Ond{\v{r}}ej Bojar, Marta~R. Costa-juss{\`a}, Christian Federmann, Yvette Graham, Roman Grundkiewicz, Barry Haddow, Matthias Huck, Eric Joanis, Tom Kocmi, Philipp Koehn, Chi-kiu Lo, Nikola Ljube{\v{s}}i{\'c}, Christof Monz, Makoto Morishita, Masaaki Nagata, Toshiaki Nakazawa, Santanu Pal, Matt Post, and Marcos Zampieri. 2020.
\newblock \href {https://aclanthology.org/2020.wmt-1.1} {Findings of the 2020 conference on machine translation ({WMT}20)}.
\newblock In \emph{Proceedings of the Fifth Conference on Machine Translation}, pages 1--55, Online. Association for Computational Linguistics.

\bibitem[{Brown et~al.(2020)Brown, Mann, Ryder, Subbiah, Kaplan, Dhariwal, Neelakantan, Shyam, Sastry, Askell et~al.}]{gpt3_brown2020language}
Tom Brown, Benjamin Mann, Nick Ryder, Melanie Subbiah, Jared~D Kaplan, Prafulla Dhariwal, Arvind Neelakantan, Pranav Shyam, Girish Sastry, Amanda Askell, et~al. 2020.
\newblock Language models are few-shot learners.
\newblock \emph{Advances in neural information processing systems}, 33:1877--1901.

\bibitem[{Chowdhery et~al.(2022)Chowdhery, Narang, Devlin, Bosma, Mishra, Roberts, Barham, Chung, Sutton, Gehrmann et~al.}]{palm_chowdhery2022palm}
Aakanksha Chowdhery, Sharan Narang, Jacob Devlin, Maarten Bosma, Gaurav Mishra, Adam Roberts, Paul Barham, Hyung~Won Chung, Charles Sutton, Sebastian Gehrmann, et~al. 2022.
\newblock Palm: Scaling language modeling with pathways.
\newblock \emph{arXiv preprint arXiv:2204.02311}.

\bibitem[{Clark et~al.(2020)Clark, Choi, Collins, Garrette, Kwiatkowski, Nikolaev, and Palomaki}]{tqdiqa_clark2020tydi}
Jonathan~H Clark, Eunsol Choi, Michael Collins, Dan Garrette, Tom Kwiatkowski, Vitaly Nikolaev, and Jennimaria Palomaki. 2020.
\newblock Tydi qa: A benchmark for information-seeking question answering in ty pologically di verse languages.
\newblock \emph{Transactions of the Association for Computational Linguistics}, 8:454--470.

\bibitem[{Conneau et~al.(2020)Conneau, Khandelwal, Goyal, Chaudhary, Wenzek, Guzm{\'a}n, Grave, Ott, Zettlemoyer, and Stoyanov}]{xlmr}
Alexis Conneau, Kartikay Khandelwal, Naman Goyal, Vishrav Chaudhary, Guillaume Wenzek, Francisco Guzm{\'a}n, Edouard Grave, Myle Ott, Luke Zettlemoyer, and Veselin Stoyanov. 2020.
\newblock \href {https://doi.org/10.18653/v1/2020.acl-main.747} {Unsupervised cross-lingual representation learning at scale}.
\newblock In \emph{Proceedings of the 58th Annual Meeting of the Association for Computational Linguistics}, pages 8440--8451, Online. Association for Computational Linguistics.

\bibitem[{Conneau and Lample(2019)}]{conneau2019cross_xlm}
Alexis Conneau and Guillaume Lample. 2019.
\newblock Cross-lingual language model pretraining.
\newblock In H.~Wallach, H.~Larochelle, A.~Beygelzimer, F.~d\' Alch\'{e}-Buc, E.~Fox, and R.~Garnett, editors, \emph{Advances in Neural Information Processing Systems 32}, pages 7059--7069. Curran Associates, Inc.

\bibitem[{Costa-juss{\`a} et~al.(2022)Costa-juss{\`a}, Cross, {\c{C}}elebi, Elbayad, Heafield, Heffernan, Kalbassi, Lam, Licht, Maillard et~al.}]{nllb_costa2022no_flores200}
Marta~R Costa-juss{\`a}, James Cross, Onur {\c{C}}elebi, Maha Elbayad, Kenneth Heafield, Kevin Heffernan, Elahe Kalbassi, Janice Lam, Daniel Licht, Jean Maillard, et~al. 2022.
\newblock No language left behind: Scaling human-centered machine translation.
\newblock \emph{arXiv preprint arXiv:2207.04672}.

\bibitem[{Dai et~al.(2022)Dai, Sun, Dong, Hao, Sui, and Wei}]{icl_gradient_descen_dai2022can}
Damai Dai, Yutao Sun, Li~Dong, Yaru Hao, Zhifang Sui, and Furu Wei. 2022.
\newblock Why can gpt learn in-context? language models secretly perform gradient descent as meta optimizers.
\newblock \emph{arXiv preprint arXiv:2212.10559}.

\bibitem[{Edunov et~al.(2018)Edunov, Ott, Auli, and Grangier}]{understanding_backtranslation_scale}
Sergey Edunov, Myle Ott, Michael Auli, and David Grangier. 2018.
\newblock \href {https://doi.org/10.18653/v1/D18-1045} {Understanding back-translation at scale}.
\newblock In \emph{Proceedings of the 2018 Conference on Empirical Methods in Natural Language Processing}, pages 489--500, Brussels, Belgium. Association for Computational Linguistics.

\bibitem[{Emezue and Dossou(2021)}]{mmtafrica-emezue-dossou-2021}
Chris~Chinenye Emezue and Bonaventure F.~P. Dossou. 2021.
\newblock \href {https://aclanthology.org/2021.wmt-1.48} {{MMTA}frica: Multilingual machine translation for {A}frican languages}.
\newblock In \emph{Proceedings of the Sixth Conference on Machine Translation}, pages 398--411, Online. Association for Computational Linguistics.

\bibitem[{Garcia et~al.(2020)Garcia, Foret, Sellam, and Parikh}]{multilingual_view_garcia-etal-2020}
Xavier Garcia, Pierre Foret, Thibault Sellam, and Ankur Parikh. 2020.
\newblock \href {https://doi.org/10.18653/v1/2020.findings-emnlp.283} {A multilingual view of unsupervised machine translation}.
\newblock In \emph{Findings of the Association for Computational Linguistics: EMNLP 2020}, pages 3160--3170, Online. Association for Computational Linguistics.

\bibitem[{Garcia et~al.(2021)Garcia, Siddhant, Firat, and Parikh}]{harness-multilingual-garcia-etal-2021}
Xavier Garcia, Aditya Siddhant, Orhan Firat, and Ankur Parikh. 2021.
\newblock \href {https://doi.org/10.18653/v1/2021.naacl-main.89} {Harnessing multilinguality in unsupervised machine translation for rare languages}.
\newblock In \emph{Proceedings of the 2021 Conference of the North American Chapter of the Association for Computational Linguistics: Human Language Technologies}, pages 1126--1137, Online. Association for Computational Linguistics.

\bibitem[{Goyal et~al.(2022)Goyal, Gao, Chaudhary, Chen, Wenzek, Ju, Krishnan, Ranzato, Guzm{\'a}n, and Fan}]{flores101-goyal-etal-2022}
Naman Goyal, Cynthia Gao, Vishrav Chaudhary, Peng-Jen Chen, Guillaume Wenzek, Da~Ju, Sanjana Krishnan, Marc{'}Aurelio Ranzato, Francisco Guzm{\'a}n, and Angela Fan. 2022.
\newblock \href {https://doi.org/10.1162/tacl_a_00474} {The {F}lores-101 evaluation benchmark for low-resource and multilingual machine translation}.
\newblock \emph{Transactions of the Association for Computational Linguistics}, 10:522--538.

\bibitem[{Guzm{\'a}n et~al.(2019)Guzm{\'a}n, Chen, Ott, Pino, Lample, Koehn, Chaudhary, and Ranzato}]{flores}
Francisco Guzm{\'a}n, Peng-Jen Chen, Myle Ott, Juan Pino, Guillaume Lample, Philipp Koehn, Vishrav Chaudhary, and Marc{'}Aurelio Ranzato. 2019.
\newblock \href {https://doi.org/10.18653/v1/D19-1632} {The {FLORES} evaluation datasets for low-resource machine translation: {N}epali{--}{E}nglish and {S}inhala{--}{E}nglish}.
\newblock In \emph{Proceedings of the 2019 Conference on Empirical Methods in Natural Language Processing and the 9th International Joint Conference on Natural Language Processing (EMNLP-IJCNLP)}, pages 6098--6111, Hong Kong, China. Association for Computational Linguistics.

\bibitem[{Hangya et~al.(2022)Hangya, Saadi, and Fraser}]{improve_low_res_mllm}
Viktor Hangya, Hossain~Shaikh Saadi, and Alexander Fraser. 2022.
\newblock \href {https://doi.org/10.18653/v1/2022.emnlp-main.822} {Improving low-resource languages in pre-trained multilingual language models}.
\newblock In \emph{Proceedings of the 2022 Conference on Empirical Methods in Natural Language Processing}, pages 11993--12006, Abu Dhabi, United Arab Emirates. Association for Computational Linguistics.

\bibitem[{Hendy et~al.(2023)Hendy, Abdelrehim, Sharaf, Raunak, Gabr, Matsushita, Kim, Afify, and Awadalla}]{howgood_gpt_mt_hendy2023good}
Amr Hendy, Mohamed Abdelrehim, Amr Sharaf, Vikas Raunak, Mohamed Gabr, Hitokazu Matsushita, Young~Jin Kim, Mohamed Afify, and Hany~Hassan Awadalla. 2023.
\newblock How good are gpt models at machine translation? a comprehensive evaluation.
\newblock \emph{arXiv preprint arXiv:2302.09210}.

\bibitem[{Hu et~al.(2021)Hu, Shen, Wallis, Allen-Zhu, Li, Wang, Wang, and Chen}]{lora_hu2021lora}
Edward~J Hu, Yelong Shen, Phillip Wallis, Zeyuan Allen-Zhu, Yuanzhi Li, Shean Wang, Lu~Wang, and Weizhu Chen. 2021.
\newblock Lora: Low-rank adaptation of large language models.
\newblock \emph{arXiv preprint arXiv:2106.09685}.

\bibitem[{Huang et~al.(2023)Huang, Tang, Zhang, Zhao, Song, Xia, and Wei}]{xlt_not_all_created_equal_huang2023not}
Haoyang Huang, Tianyi Tang, Dongdong Zhang, Wayne~Xin Zhao, Ting Song, Yan Xia, and Furu Wei. 2023.
\newblock Not all languages are created equal in llms: Improving multilingual capability by cross-lingual-thought prompting.
\newblock \emph{arXiv preprint arXiv:2305.07004}.

\bibitem[{Lample et~al.(2018)Lample, Ott, Conneau, Denoyer, and Ranzato}]{lample2018phrase_unsup}
Guillaume Lample, Myle Ott, Alexis Conneau, Ludovic Denoyer, and Marc{'}Aurelio Ranzato. 2018.
\newblock \href {https://doi.org/10.18653/v1/D18-1549} {Phrase-based {\&} neural unsupervised machine translation}.
\newblock In \emph{Proceedings of the 2018 Conference on Empirical Methods in Natural Language Processing}, pages 5039--5049, Brussels, Belgium. Association for Computational Linguistics.

\bibitem[{Lauren{\c{c}}on et~al.(2022)Lauren{\c{c}}on, Saulnier, Wang, Akiki, del Moral, Scao, Werra, Mou, Ponferrada, Nguyen, Frohberg, {\v{S}}a{\v{s}}ko, Lhoest, McMillan-Major, Dupont, Biderman, Rogers, allal, Toni, Pistilli, Nguyen, Nikpoor, Masoud, Colombo, de~la Rosa, Villegas, Thrush, Longpre, Nagel, Weber, Mu{\~n}oz, Zhu, Strien, Alyafeai, Almubarak, Chien, Gonzalez-Dios, Soroa, Lo, Dey, Suarez, Gokaslan, Bose, Adelani, Phan, Tran, Yu, Pai, Chim, Lepercq, Ilic, Mitchell, Luccioni, and Jernite}]{roots_corpus}
Hugo Lauren{\c{c}}on, Lucile Saulnier, Thomas Wang, Christopher Akiki, Albert~Villanova del Moral, Teven~Le Scao, Leandro~Von Werra, Chenghao Mou, Eduardo~Gonz{\'a}lez Ponferrada, Huu Nguyen, J{\"o}rg Frohberg, Mario {\v{S}}a{\v{s}}ko, Quentin Lhoest, Angelina McMillan-Major, G{\'e}rard Dupont, Stella Biderman, Anna Rogers, Loubna~Ben allal, Francesco~De Toni, Giada Pistilli, Olivier Nguyen, Somaieh Nikpoor, Maraim Masoud, Pierre Colombo, Javier de~la Rosa, Paulo Villegas, Tristan Thrush, Shayne Longpre, Sebastian Nagel, Leon Weber, Manuel~Romero Mu{\~n}oz, Jian Zhu, Daniel~Van Strien, Zaid Alyafeai, Khalid Almubarak, Vu~Minh Chien, Itziar Gonzalez-Dios, Aitor Soroa, Kyle Lo, Manan Dey, Pedro~Ortiz Suarez, Aaron Gokaslan, Shamik Bose, David~Ifeoluwa Adelani, Long Phan, Hieu Tran, Ian Yu, Suhas Pai, Jenny Chim, Violette Lepercq, Suzana Ilic, Margaret Mitchell, Sasha Luccioni, and Yacine Jernite. 2022.
\newblock \href {https://openreview.net/forum?id=UoEw6KigkUn} {The bigscience {ROOTS} corpus: A 1.6{TB} composite multilingual dataset}.
\newblock In \emph{Thirty-sixth Conference on Neural Information Processing Systems Datasets and Benchmarks Track}.

\bibitem[{Limisiewicz et~al.(2023)Limisiewicz, Balhar, and Mare{\v{c}}ek}]{tokenization_impact_multilingual_lm}
Tomasz Limisiewicz, Ji{\v{r}}{\'\i} Balhar, and David Mare{\v{c}}ek. 2023.
\newblock \href {https://doi.org/10.18653/v1/2023.findings-acl.350} {Tokenization impacts multilingual language modeling: Assessing vocabulary allocation and overlap across languages}.
\newblock In \emph{Findings of the Association for Computational Linguistics: ACL 2023}, pages 5661--5681, Toronto, Canada. Association for Computational Linguistics.

\bibitem[{Lin(2004)}]{lin2004rouge}
Chin-Yew Lin. 2004.
\newblock Rouge: A package for automatic evaluation of summaries.
\newblock In \emph{Text summarization branches out}, pages 74--81.

\bibitem[{Lin et~al.(2022)Lin, Mihaylov, Artetxe, Wang, Chen, Simig, Ott, Goyal, Bhosale, Du, Pasunuru, Shleifer, Koura, Chaudhary, O{'}Horo, Wang, Zettlemoyer, Kozareva, Diab, Stoyanov, and Li}]{few_shot_multilingual_llm}
Xi~Victoria Lin, Todor Mihaylov, Mikel Artetxe, Tianlu Wang, Shuohui Chen, Daniel Simig, Myle Ott, Naman Goyal, Shruti Bhosale, Jingfei Du, Ramakanth Pasunuru, Sam Shleifer, Punit~Singh Koura, Vishrav Chaudhary, Brian O{'}Horo, Jeff Wang, Luke Zettlemoyer, Zornitsa Kozareva, Mona Diab, Veselin Stoyanov, and Xian Li. 2022.
\newblock \href {https://doi.org/10.18653/v1/2022.emnlp-main.616} {Few-shot learning with multilingual generative language models}.
\newblock In \emph{Proceedings of the 2022 Conference on Empirical Methods in Natural Language Processing}, pages 9019--9052, Abu Dhabi, United Arab Emirates. Association for Computational Linguistics.

\bibitem[{Liu et~al.(2023)Liu, Iter, Xu, Wang, Xu, and Zhu}]{gpt_eval_liu2023g-eval}
Yang Liu, Dan Iter, Yichong Xu, Shuohang Wang, Ruochen Xu, and Chenguang Zhu. 2023.
\newblock \href {https://www.microsoft.com/en-us/research/publication/gpteval-nlg-evaluation-using-gpt-4-with-better-human-alignment/} {G-eval: Nlg evaluation using gpt-4 with better human alignment}.
\newblock \emph{arXiv 2303.16634}.

\bibitem[{Liu et~al.(2020)Liu, Gu, Goyal, Li, Edunov, Ghazvininejad, Lewis, and Zettlemoyer}]{mbart_liu2020multilingual}
Yinhan Liu, Jiatao Gu, Naman Goyal, Xian Li, Sergey Edunov, Marjan Ghazvininejad, Mike Lewis, and Luke Zettlemoyer. 2020.
\newblock Multilingual denoising pre-training for neural machine translation.
\newblock \emph{arXiv preprint arXiv:2001.08210}.

\bibitem[{Min et~al.(2022)Min, Lyu, Holtzman, Artetxe, Lewis, Hajishirzi, and Zettlemoyer}]{rethink_demonstration_min2022rethinking}
Sewon Min, Xinxi Lyu, Ari Holtzman, Mikel Artetxe, Mike Lewis, Hannaneh Hajishirzi, and Luke Zettlemoyer. 2022.
\newblock Rethinking the role of demonstrations: What makes in-context learning work?
\newblock \emph{arXiv preprint arXiv:2202.12837}.

\bibitem[{Muennighoff et~al.(2022)Muennighoff, Wang, Sutawika, Roberts, Biderman, Scao, Bari, Shen, Yong, Schoelkopf et~al.}]{bloomz_muennighoff2022crosslingual}
Niklas Muennighoff, Thomas Wang, Lintang Sutawika, Adam Roberts, Stella Biderman, Teven~Le Scao, M~Saiful Bari, Sheng Shen, Zheng-Xin Yong, Hailey Schoelkopf, et~al. 2022.
\newblock Crosslingual generalization through multitask finetuning.
\newblock \emph{arXiv preprint arXiv:2211.01786}.

\bibitem[{Narayan et~al.(2018)Narayan, Cohen, and Lapata}]{xsum_Narayan2018DontGM}
Shashi Narayan, Shay~B. Cohen, and Mirella Lapata. 2018.
\newblock Don't give me the details, just the summary! topic-aware convolutional neural networks for extreme summarization.
\newblock \emph{ArXiv}, abs/1808.08745.

\bibitem[{Nguyen et~al.(2022{\natexlab{a}})Nguyen, Gong, Tang, Wang, Koehn, and Joty}]{swavumt_nguyen2022contrastive}
Xuan-Phi Nguyen, Hongyu Gong, Yun Tang, Changhan Wang, Philipp Koehn, and Shafiq Joty. 2022{\natexlab{a}}.
\newblock \href {https://openreview.net/forum?id=pN1JOdrSY9} {Contrastive clustering to mine pseudo parallel data for unsupervised translation}.
\newblock In \emph{International Conference on Learning Representations}.

\bibitem[{Nguyen et~al.(2022{\natexlab{b}})Nguyen, Joty, Wu, and Aw}]{refine_lowres_umt_nguyen2022}
Xuan-Phi Nguyen, Shafiq Joty, Kui Wu, and AiTi Aw. 2022{\natexlab{b}}.
\newblock \href {https://openreview.net/forum?id=eCUeRHHupF} {Refining low-resource unsupervised translation by language disentanglement of multilingual translation model}.
\newblock In \emph{Advances in Neural Information Processing Systems}.

\bibitem[{Nguyen et~al.(2023)Nguyen, Zhang, Li, Aljunied, Tan, Cheng, Chen, Deng, Yang, Liu et~al.}]{seallm_nguyen2023seallms}
Xuan-Phi Nguyen, Wenxuan Zhang, Xin Li, Mahani Aljunied, Qingyu Tan, Liying Cheng, Guanzheng Chen, Yue Deng, Sen Yang, Chaoqun Liu, et~al. 2023.
\newblock Seallms--large language models for southeast asia.
\newblock \emph{arXiv preprint arXiv:2312.00738}.

\bibitem[{OpenAI(2023)}]{gpt4}
OpenAI. 2023.
\newblock Gpt-4 technical report.
\newblock \emph{arXiv preprint}.

\bibitem[{Ouyang et~al.(2022)Ouyang, Wu, Jiang, Almeida, Wainwright, Mishkin, Zhang, Agarwal, Slama, Ray et~al.}]{instructgpt_ouyang2022training}
Long Ouyang, Jeffrey Wu, Xu~Jiang, Diogo Almeida, Carroll Wainwright, Pamela Mishkin, Chong Zhang, Sandhini Agarwal, Katarina Slama, Alex Ray, et~al. 2022.
\newblock Training language models to follow instructions with human feedback.
\newblock \emph{Advances in Neural Information Processing Systems}, 35:27730--27744.

\bibitem[{Popovi{\'c}(2015)}]{popovic2015chrf}
Maja Popovi{\'c}. 2015.
\newblock chrf: character n-gram f-score for automatic mt evaluation.
\newblock In \emph{Proceedings of the tenth workshop on statistical machine translation}, pages 392--395.

\bibitem[{Post(2018)}]{sacredbleu_post-2018-call}
Matt Post. 2018.
\newblock \href {https://doi.org/10.18653/v1/W18-6319} {A call for clarity in reporting {BLEU} scores}.
\newblock In \emph{Proceedings of the Third Conference on Machine Translation: Research Papers}, pages 186--191, Brussels, Belgium. Association for Computational Linguistics.

\bibitem[{Radford et~al.(2019)Radford, Wu, Child, Luan, Amodei, and Sutskever}]{gpt2_radford2019language_gpt2}
Alec Radford, Jeffrey Wu, Rewon Child, David Luan, Dario Amodei, and Ilya Sutskever. 2019.
\newblock Language models are unsupervised multitask learners.
\newblock \emph{OpenAI Blog}, 1(8):9.

\bibitem[{Scao et~al.(2022)Scao, Fan, Akiki, Pavlick, Ili{\'c}, Hesslow, Castagn{\'e}, Luccioni, Yvon, Gall{\'e} et~al.}]{bloom_scao2022bloom}
Teven~Le Scao, Angela Fan, Christopher Akiki, Ellie Pavlick, Suzana Ili{\'c}, Daniel Hesslow, Roman Castagn{\'e}, Alexandra~Sasha Luccioni, Fran{\c{c}}ois Yvon, Matthias Gall{\'e}, et~al. 2022.
\newblock Bloom: A 176b-parameter open-access multilingual language model.
\newblock \emph{arXiv preprint arXiv:2211.05100}.

\bibitem[{Shi et~al.(2022)Shi, Suzgun, Freitag, Wang, Srivats, Vosoughi, Chung, Tay, Ruder, Zhou et~al.}]{llm_are_mling_cot_shi2022language}
Freda Shi, Mirac Suzgun, Markus Freitag, Xuezhi Wang, Suraj Srivats, Soroush Vosoughi, Hyung~Won Chung, Yi~Tay, Sebastian Ruder, Denny Zhou, et~al. 2022.
\newblock Language models are multilingual chain-of-thought reasoners.
\newblock \emph{arXiv preprint arXiv:2210.03057}.

\bibitem[{Sia and Duh(2023)}]{icl_maintain_coherence_sia2023context}
Suzanna Sia and Kevin Duh. 2023.
\newblock In-context learning as maintaining coherency: A study of on-the-fly machine translation using large language models.
\newblock \emph{arXiv preprint arXiv:2305.03573}.

\bibitem[{Tang et~al.(2020)Tang, Tran, Li, Chen, Goyal, Chaudhary, Gu, and Fan}]{mbart50_tang2020multilingual}
Yuqing Tang, Chau Tran, Xian Li, Peng-Jen Chen, Naman Goyal, Vishrav Chaudhary, Jiatao Gu, and Angela Fan. 2020.
\newblock Multilingual translation with extensible multilingual pretraining and finetuning.
\newblock \emph{arXiv preprint arXiv:2008.00401}.

\bibitem[{Touvron et~al.(2023{\natexlab{a}})Touvron, Lavril, Izacard, Martinet, Lachaux, Lacroix, Rozi{\`e}re, Goyal, Hambro, Azhar, Rodriguez, Joulin, Grave, and Lample}]{llama_touvron2023}
Hugo Touvron, Thibaut Lavril, Gautier Izacard, Xavier Martinet, Marie-Anne Lachaux, Timoth{\'e}e Lacroix, Baptiste Rozi{\`e}re, Naman Goyal, Eric Hambro, Faisal Azhar, Aurelien Rodriguez, Armand Joulin, Edouard Grave, and Guillaume Lample. 2023{\natexlab{a}}.
\newblock Llama: Open and efficient foundation language models.
\newblock \emph{arXiv preprint arXiv:2302.13971}.

\bibitem[{Touvron et~al.(2023{\natexlab{b}})Touvron, Martin, Stone, Albert, Almahairi, Babaei, Bashlykov, Batra, Bhargava, Bhosale et~al.}]{llama2_touvron2023llama}
Hugo Touvron, Louis Martin, Kevin Stone, Peter Albert, Amjad Almahairi, Yasmine Babaei, Nikolay Bashlykov, Soumya Batra, Prajjwal Bhargava, Shruti Bhosale, et~al. 2023{\natexlab{b}}.
\newblock Llama 2: Open foundation and fine-tuned chat models.
\newblock \emph{arXiv preprint arXiv:2307.09288}.

\bibitem[{Tran et~al.(2020)Tran, Tang, Li, and Gu}]{criss2020}
Chau Tran, Yuqing Tang, Xian Li, and Jiatao Gu. 2020.
\newblock \href {https://proceedings.neurips.cc/paper/2020/file/1763ea5a7e72dd7ee64073c2dda7a7a8-Paper.pdf} {Cross-lingual retrieval for iterative self-supervised training}.
\newblock In \emph{Advances in Neural Information Processing Systems}, volume~33, pages 2207--2219. Curran Associates, Inc.

\bibitem[{{\"U}st{\"u}n et~al.(2024){\"U}st{\"u}n, Aryabumi, Yong, Ko, D'souza, Onilude, Bhandari, Singh, Ooi, Kayid et~al.}]{aya_model}
Ahmet {\"U}st{\"u}n, Viraat Aryabumi, Zheng-Xin Yong, Wei-Yin Ko, Daniel D'souza, Gbemileke Onilude, Neel Bhandari, Shivalika Singh, Hui-Lee Ooi, Amr Kayid, et~al. 2024.
\newblock Aya model: An instruction finetuned open-access multilingual language model.
\newblock \emph{arXiv preprint arXiv:2402.07827}.

\bibitem[{Wei et~al.(2022)Wei, Wang, Schuurmans, Bosma, Chi, Le, and Zhou}]{cot_wei2022chain}
Jason Wei, Xuezhi Wang, Dale Schuurmans, Maarten Bosma, Ed~Chi, Quoc Le, and Denny Zhou. 2022.
\newblock Chain of thought prompting elicits reasoning in large language models.
\newblock \emph{arXiv preprint arXiv:2201.11903}.

\bibitem[{Wei et~al.(2023)Wei, Wei, Tay, Tran, Webson, Lu, Chen, Liu, Huang, Zhou et~al.}]{llm_do_icl_differently_wei2023larger}
Jerry Wei, Jason Wei, Yi~Tay, Dustin Tran, Albert Webson, Yifeng Lu, Xinyun Chen, Hanxiao Liu, Da~Huang, Denny Zhou, et~al. 2023.
\newblock Larger language models do in-context learning differently.
\newblock \emph{arXiv preprint arXiv:2303.03846}.

\bibitem[{Wenzek et~al.(2020)Wenzek, Lachaux, Conneau, Chaudhary, Guzm{\'a}n, Joulin, and Grave}]{ccnet_wenzek-etal-2020-ccnet}
Guillaume Wenzek, Marie-Anne Lachaux, Alexis Conneau, Vishrav Chaudhary, Francisco Guzm{\'a}n, Armand Joulin, and Edouard Grave. 2020.
\newblock \href {https://www.aclweb.org/anthology/2020.lrec-1.494} {{CCN}et: Extracting high quality monolingual datasets from web crawl data}.
\newblock In \emph{Proceedings of the 12th Language Resources and Evaluation Conference}, pages 4003--4012, Marseille, France. European Language Resources Association.

\bibitem[{Xie et~al.(2021)Xie, Raghunathan, Liang, and Ma}]{icl_bayesian_infer_xie2021explanation}
Sang~Michael Xie, Aditi Raghunathan, Percy Liang, and Tengyu Ma. 2021.
\newblock An explanation of in-context learning as implicit bayesian inference.
\newblock \emph{arXiv preprint arXiv:2111.02080}.

\bibitem[{Zhang et~al.(2023)Zhang, Haddow, and Birch}]{prompt_llm_mt_casestudy_zhang2023prompting}
Biao Zhang, Barry Haddow, and Alexandra Birch. 2023.
\newblock Prompting large language model for machine translation: A case study.
\newblock \emph{arXiv preprint arXiv:2301.07069}.

\bibitem[{Zheng et~al.(2023)Zheng, Chiang, Sheng, Zhuang, Wu, Zhuang, Lin, Li, Li, Xing et~al.}]{mt_bench_zheng2023judging}
Lianmin Zheng, Wei-Lin Chiang, Ying Sheng, Siyuan Zhuang, Zhanghao Wu, Yonghao Zhuang, Zi~Lin, Zhuohan Li, Dacheng Li, Eric Xing, et~al. 2023.
\newblock Judging llm-as-a-judge with mt-bench and chatbot arena.
\newblock \emph{arXiv preprint arXiv:2306.05685}.

\bibitem[{Zhou et~al.(2023)Zhou, Liu, Xu, Iyer, Sun, Mao, Ma, Efrat, Yu, Yu et~al.}]{lima_zhou2023lima}
Chunting Zhou, Pengfei Liu, Puxin Xu, Srini Iyer, Jiao Sun, Yuning Mao, Xuezhe Ma, Avia Efrat, Ping Yu, Lili Yu, et~al. 2023.
\newblock Lima: Less is more for alignment.
\newblock \emph{arXiv preprint arXiv:2305.11206}.

\bibitem[{Zhu et~al.(2023)Zhu, Liu, Dong, Xu, Kong, Chen, Li, and Huang}]{mmt_llm_analysis_zhu2023multilingual}
Wenhao Zhu, Hongyi Liu, Qingxiu Dong, Jingjing Xu, Lingpeng Kong, Jiajun Chen, Lei Li, and Shujian Huang. 2023.
\newblock Multilingual machine translation with large language models: Empirical results and analysis.
\newblock \emph{arXiv preprint arXiv:2304.04675}.

\end{thebibliography}
\bibliographystyle{acl_natbib}

\appendix

\section{Example Appendix}

\subsection{Low-resource Language Details}\label{subsec:lang_coverage_test}

\Cref{table:lang_details} lists the details of each low-resource language in the ROOTS corpus \citep{roots_corpus} that we mainly evaluate with the BLOOM model \citep{bloom_scao2022bloom}. Regarding test sets, we primarily choose from the ML50 benchmark \citep{mbart50_tang2020multilingual}, which collected test data from various sources, such as WMT \citep{wmt2020_barrault-etal-2020-findings} and FLoRes \citep{flores,flores101-goyal-etal-2022}. For languages absent in ML50, we choose the NLLB-devtest sets \citep{nllb_costa2022no_flores200} as replacement. For non-English $X$$\rightarrow$$Y$ tasks, we choose NLLB-devtest for all our evaluation. To limit the API call costs within our budget, we randomly the same 200 samples from each test set for evaluation.

\begin{table*}[h]
    \centering
    \begin{tabular}{lccc|lccc}
    \hline
    \multicolumn{4}{c|}{\bf Indic}  & \multicolumn{4}{c}{\bf African}  \\
    {\bf Name} & {\bf Code} & {\bf Test} & {\bf Unlabeled} & {\bf Name} & {\bf Code} & {\bf Test set} & {\bf Unlabeled} \\
    Assamese	& as	& NLLB	 & CC100       & Tumbuka         & --/tum  & NLLB & OUR \\
    Oriya		& or	& NLLB	 & ROOTS       & Kikuyu          & ki/kik  & NLLB & OUR \\
    Gujarati	& gu	& ML50	 & CC100       & Bambara         & bm/bam  & NLLB & MAFAND \\
    Marathi		& mr	& ML50	 & CC100       & Akan            & ak/aka  & NLLB & OUR \\
    Panjabi		& pa	& NLLB	 & CC100       & Tsonga          & ts/tso  & NLLB & MMTAfrica \\
    Kannada		& kn	& NLLB	 & CC100       & Southern Sotho  & st/sot  & NLLB & OUR \\
    Nepali		& ne	& ML50	 & CC100       & Chewa           & ny/nya  & NLLB & MMTAfrica \\
    Telugu		& te	& ML50	 & CC100       & Tswana          & tn/tsn  & NLLB & MMTAfrica \\
    Malayalam	& ml	& ML50	 & CC100       & Lingala         & ln/lin  & NLLB & MMTAfrica \\
    Urdu		& ur	& NLLB	 & CC100       & Northern Sotho  & --/nso  & NLLB & MMTAfrica \\
    Tamil		& ta	& ML50	 & CC100       & Fon             & --/fon  & NLLB & MAFAND \\
    Bengali		& bn	& NLLB	 & CC100       & Rundi           & rn/run  & NLLB & OUR \\
    Hindi		& hi	& ML50	 & CC100       & Wolof           & wo/wol  & NLLB & CC100 \\
    & &                          & CC100       & Luganda         & lg/lug  & NLLB & CC100 \\
    & &                          & CC100       & Shona           & sn/sna  & NLLB & CC100 \\
    & &                          & CC100       & Zulu            & zu/zul  & NLLB & CC100 \\
    & &                          & CC100       & Igbo            & ig/ibo  & NLLB & CC100 \\
    & &                          & CC100       & Xhosa           & xh/xho  & NLLB & CC100 \\
    & &                          & CC100       & Kinyarwanda     & rw/kin  & NLLB & MMTAfrica \\
    & &                          & CC100       & Yoruba          & yo/yor  & NLLB & CC100 \\
    & &                          & CC100       & Swahili         & sw/swa  & NLLB & CC100 \\
    \hline
    \end{tabular}
    \caption{
      Low-resource language details and corresponding test sets and unlabeled data sources for $X$$\leftrightarrow$En translation tasks.
    }
    \label{table:lang_details}
\end{table*}


\begin{figure*}[h]
  \centering
  \begin{tikzpicture}
  \begin{axis}[
    col bar axis style,
    xticklabels={,,},
    xtick=\empty,
    ymin=0.00001, 
    ymax=1,
    ybar=0.1em,
    bar width=3pt,
    width=\textwidth,
    ymode=log,
    log origin y=infty,
  ]
    \addplot table[x=id, y=percent ] \langpercentafrican;
    \addplot table[x=id, y=percent ] \langpercentindic;
    \legend{African, Indic}
  \end{axis}
  \end{tikzpicture}
  \caption{Low-resource language coverage \% of the ROOTS corpus \citep{roots_corpus} used to train BLOOM. The highest-resource language for Indic and African are Hindi and Swahili. Hindi accounts for $0.7$\% and the rarest language, Tumbuka, takes up only $2e^{-5}$\% of the corpus.}
  \label{fig:bloom_lang_coverage}
\end{figure*}

\subsection{Experiment Details}\label{subsec:exp_details}



\paragraph{Few-shot data sources.}For supervised prompting, we collect randomly parallel pairs from the respective valid set for each language. For unlabeled data for our LDP method, we collect and filter data from various sources, as specified in \textit{Unlabeled} column of \Cref{table:lang_details}. Specifically, the primary unlabeled source is the CC100 corpus \citep{ccnet_wenzek-etal-2020-ccnet,xlmr}. For those absent in CC100, we collect data from other sources, such as the ROOTS corpus \citep{roots_corpus}, MMTAfrica \citep{mmtafrica-emezue-dossou-2021} and MAFAND \citep{mafandmt_adelani-etal-2022-thousand}. For the remaining languages where we could not find in research repositories, we crawled from several religious and news websites (OUR). The sizes of collected unlabeled texts vary greatly, ranging from a few millions lines for Hindi to less than 1000 lines for Bambara, thus presenting a challenge for data balancing. For LDP non-English high-resource exemplars, we randomly collect a single high-quality sentence of similar lengths from the CC100 corpus for each language and use the unsupervised CRISS model \citep{criss2020} to translate them into English.

\paragraph{Unlabeled data filtering}To ensure high-quality native texts for unsupervised LDP prompting as well as larger-scale synthetic data creation for fine-tuning, we filter unlabeled texts such that they \Ni are within 20 to 200 character lengths, \Nii do not contain non-conversational artifacts like URLs, brackets, bullet points or excessive numbers, and \Niii do not contain more than 20\% alphabetical characters for Indic and non-latin characters for African languages. For fine-tuning, we use an upscaling temperature of 25 to smoothen the data mixture imbalance.

\paragraph{Other Details.}We evaluate translation tasks with chrF++ \citep{popovic2015chrf} and SacreBLEU \citep{sacredbleu_post-2018-call}. For SacreBLEU, we use the default tokenizer for Latin-based languages, while follow \citet{flores,flores101-goyal-etal-2022} to use \href{https://github.com/anoopkunchukuttan/indic_nlp_library}{indic\_nlp\_library} for Indic language tokenization.

For each of the 68 language pairs, we sample randomly and evaluate the same 200 sentences from each test set with the same zero seed to limit the cost of API calls\footnote{\href{https://huggingface.co/bigscience/bloom}{bigscience/bloom}, \href{https://platform.openai.com}{openai.com}.}. We conduct full-set evaluations for 4 random languages in each group and observe $<1$ chrF++ standard deviation from our 200-sample evaluations.

\begin{table}[h]
  \centering
  \resizebox{\columnwidth}{!}{%
  \begin{tabular}{l|cccc}
    \hline
    \multirow{2}{*}{\bf LLaMA-30B}  & \multicolumn{2}{c}{\bf X$\rightarrow$En}  & \multicolumn{2}{c}{\bf En$\rightarrow$X}  \\
    \cmidrule(lr){2-5} 
    & {\bf chrF++} & {\bf BLEU}  & {\bf chrF++} & {\bf BLEU}   \\
    \hline
    Supervised        & 61.80	& 39.51	& 53.65	& 28.98 \\
    Unsupervised-LDP  & 61.75	& 38.83	& 54.00	& 29.58 \\
    \hline
  \end{tabular}
  }
  \caption{
    Comparison between supervised and unsupervised-LDP prompting with LLaMA-30B model in translation tasks between English (En) and 19 European languages (X). 
    LDP prompts consist of exemplars from high-resource languages seen by CRISS.
  }
  \label{table:unsup_llama}
\end{table}

\paragraph{Low-resource $\leftrightarrow$ En translation.}

add something here

\begin{table*}[h]
  \centering
  \resizebox{\textwidth}{!}{%
  \begin{tabular}{l|cccccccc}
  \hline
    & \multicolumn{2}{c}{\bf Ind-En}  & \multicolumn{2}{c}{\bf En-Ind}  & \multicolumn{2}{c}{\bf Afr-En}  & \multicolumn{2}{c}{\bf En-Afr} \\
    \cmidrule(lr){2-9} 
    & {\bf chrF++} & {\bf BLEU}  & {\bf chrF++} & {\bf BLEU}  & {\bf chrF++} & {\bf BLEU}  & {\bf chrF++} & {\bf BLEU}  \\
  \hline
  \multicolumn{9}{l}{\bf Foundation BLOOM-175B	}\\
  Supervised-8-shot                    &  47.31 & 22.32 & 34.66 & 9.02 & 28.64 & 8.35 & 14.93 & 2.00 \\
  Unsupervised-LDP                     &  47.62 & 22.38 & 34.54 & 8.88 & 28.72 & 8.71 & 14.57 & 1.89 \\
  \hline
  \multicolumn{9}{l}{\bf Foundation BLOOM-7B	}\\
  Supervised-8-shot                    &  39.86 & 14.77 & 24.02 & 4.42  & 21.51 & 4.33  & 11.27 & 0.59 \\
  Unsupervised-LDP                     &  39.88 & 14.96 & 24.41 & 4.52  & 20.47 & 3.65  & 12.04 & 0.62 \\
  Fine-tune QKV (2B params)             &  42.19 & 17.13 & 32.72 & 8.33  & 21.14 & 5.15  & 15.73 & 2.13 \\ 
  \hline
  \multicolumn{9}{l}{\bf Supervised RLHF InstructGPT (text-davinci-003)} \\
  Zero-shot with instruction           & 35.37 & 11.48 & 20.71 & 3.88  & 27.10 & 8.04  & 15.45 & 1.13 \\
  Supervised-6-shot                    & 37.07 & 13.13 & 24.74 & 5.21  & 31.51 & 10.88 & 19.22 & 2.66 \\
  Unsupervised-LDP                     & 38.45 & 14.22 & 25.17 & 5.06  & 31.92 & 11.12 & 19.51 & 2.61 \\
  \hline
  \multicolumn{9}{l}{\bf Supervised upperbound}\\
  NLLB-200 distilled                   &  \textit{61.00} & \textit{37.24} & \textit{46.77} & \textit{18.78} & \textit{48.42} & \textit{26.92} & \textit{39.18} & \textit{12.95} \\
  \hline
  \end{tabular}
  }
  \caption{
    Averaged performances of different prompting techniques across various model sizes and types, namely BLOOM \citep{bloom_scao2022bloom} and InstructGPT text-davinci-003 \citep{gpt3_brown2020language,instructgpt_ouyang2022training}, in translation tasks between English (En) and 13 Indic (Ind) and 21 African (Afr) low-resource languages present in the ROOTS corpus \citep{roots_corpus}.
  }
  \label{table:main_unsup_llmmt_sacrebleu}
\end{table*}








\subsection{Additional Experiments}\label{subsec:additional_exps}

\begin{figure*}[h]
    \centering
    \begin{tikzpicture}
    \begin{axis}[
      bar axis style,
      ylabel=chrF++,
      xticklabels={,,},
      xtick=\empty,
      ymin=18, 
      ymax=58,
      bar width=2pt,
      legend style={
        at={(0.3, -0.05)},
      },
    ]
      \fill[gray, opacity=0.1] ({rel axis cs:0.61,0}) rectangle ({rel axis cs:1,1});
      \addplot table[x=id, y=sup_toen ] \dataafricanindic;
      \addplot table[x=id, y=our_toen ] \dataafricanindic;
      \draw [dashed, violet, thick] (21.5,0) -- (21.5,59);
      \node[] at (axis cs: 10,55) {African};
      \node[] at (axis cs: 28,55) {Indic};
      \legend{Supervised, Unsupervised-LDP}
    \end{axis}
    \begin{axis}[
      line axis style,
      ymode=log,
      ymin=0.00001,
      ymax=1,
      log origin y=infty,
      legend style={
        at={(0.75, -0.05)},
      }
    ]
      \addplot [black, mark=*] table[x=id,y=percent] \dataafricanindic;;
      \legend{\% pre-training data}
    \end{axis}
    \end{tikzpicture}
    \caption{chrF++ scores for translation from each Indic and African language in the ROOTS corpus to English ($X$$\rightarrow$En), using BLOOM. 
    The right y-axis indicates corresponding pre-training coverage of each language at log scale.}
    \label{fig:all_prompt_scores_toen}
\end{figure*}

\begin{figure*}[h]
  \centering
  \begin{tikzpicture}
  \begin{axis}[
    bar axis style,
    ylabel=chrF++,
    xticklabels={,,},
    xtick=\empty,
    ymin=3, 
    ymax=50,
    bar width=2pt,
    legend style={
      at={(0.3, -0.05)},
    },
  ]
    \fill[gray, opacity=0.1] ({rel axis cs:0.61,0}) rectangle ({rel axis cs:1,1});
    \addplot table[x=id, y=sup_fromen ] \dataafricanindic;
    \addplot table[x=id, y=our_fromen ] \dataafricanindic;
    \draw [dashed, violet, thick] (21.5,0) -- (21.5,50);
    \node[] at (axis cs: 10,45) {African};
    \node[] at (axis cs: 28,45) {Indic};
    \legend{Supervised, Unsupervised-LDP}
  \end{axis}
  \begin{axis}[
    line axis style,
    ymode=log,
    ymin=0.00001,
    ymax=1,
    log origin y=infty,
    legend style={
      at={(0.75, -0.05)},
    }
  ]
    \addplot [black, mark=*] table[x=id,y=percent] \dataafricanindic;;
    \legend{\% pre-training data}
  \end{axis}
  \end{tikzpicture}
  \caption{chrF++ scores for translation from English to each Indic and African language in the ROOTS corpus (En$\rightarrow$$X$), using BLOOM. 
  The right y-axis indicates corresponding pre-training coverage of each language at log scale.}
  \label{fig:all_prompt_scores_tox}
\end{figure*}

\paragraph{Breakdown of $X$$\rightarrow$En.}Similar to the observation for En$\rightarrow$$X$ in the main paper, \Cref{fig:all_prompt_scores_toen} shows that LDP performs generally on par with supervised prompting equally across all languages, and that it does not unevenly perform much worse or better in any particular language.


\paragraph{High-resource Translation with Llama}LLaMA \citep{llama_touvron2023} is another open-sourced LLM that only supports 20 European high-resource languages. We evaluate LLaMA in translation tasks between English and the remaining 19 languages, which include Hungarian, Danish and Catalan. Specifically, we use CRISS to generate synthetic LDP exemplars from De, Es and Fr, which we then use to prompt LLaMA to translate from and to such languages. As reported in \Cref{table:unsup_llama}, we observe similar trends where our LDP method performs competitively with supervised prompting. The overall scores for such languages are also much higher than those of non-Latin languages because LLaMA was also pre-trained with bitexts, though without explicit alignments.

\begin{figure*}[h]
    \centering
    \begin{tikzpicture}
    \begin{axis}[
      bar axis style,
      ylabel=X/En Token ratio,
      xticklabels={,,},
      xtick=\empty,
      ymin=1, 
      ymax=17,
      bar width=2pt,
      legend style={
        at={(0.3, -0.05)},
      },
    ]
      \fill[gray, opacity=0.1] ({rel axis cs:0.61,0}) rectangle ({rel axis cs:1,1});
      \addplot table[x=id, y=bloom_tok_ratio ] \dataafricanindic;
      \addplot table[x=id, y=openai_tok_ratio ] \dataafricanindic;
      \draw [dashed, violet, thick] (21.5,0) -- (21.5,59);
      \node[] at (axis cs: 10,15.5) {African - GPT wins};
      \node[] at (axis cs: 26,15.5) {Indic - GPT loses};
      \legend{BLOOM, InstructGPT}
    \end{axis}
    \begin{axis}[
      line axis style,
      ymin=-20,
      ymax=20,
      legend style={
        at={(0.75, -0.05)},
      }
    ]
      \addplot [black, mark=*] table[x=id,y=openai_m_blo_fromen] \dataafricanindic;;
      \draw [dashed, black, thick] (-1,0) -- (36,0);
      \legend{chrF++ difference for En$\rightarrow$$X$}
    \end{axis}
    \end{tikzpicture}
    \caption{Tokenization issue. {Left y-axis bar chart}: The average ratios between the token lengths of $X$-language text over their English counterparts of the same meaning. 
    {Right y-axis line chart}: chrF++ performance difference between GPT text-davinci-003 and BLOOM for En$\rightarrow$$X$ tasks, meaning $<0$ indicates GPT is worse than BLOOM.
    }
    \label{fig:tokenization_issue_with_diff}
\end{figure*}

\paragraph{BLOOM vs. InstructGPT.}While much evidence show that InstructGPT text-davinci-003 is more superior than the vanilla BLOOM \citep{bloom_scao2022bloom,instructgpt_ouyang2022training} in many languages, our experiments with low-resource languages demonstrate it is not always true for low-resource non-Latin languages, as shown in the main paper. \Cref{fig:tokenization_issue_with_diff} explains clearly the reason is that GPT's tokenizer is not designed to allocate meaningful sub-word tokens for non-Latin texts, such as Indic lexical items, while significantly favors Latin characters due to the sheer size of Latin texts in its pre-training data. For example of InstructGPT, a 10-token English text can be equivalent to a 160-token Tamil text but only a 28-token Tumbuka text, despite Tumbuka is much more low-resource. This issue is non-existent in BLOOM, as the ratios naturally decrease when data coverages increase. As shown in the table, InstructGPT becomes worse than BLOOM as soon as the ratio between token lengths of target language over English surpass 5 in Indic languages. We refer to this as sub-word token fragmentation, where texts are broken into very long byte-level tokens that exceed the context length and suppress performances.


\paragraph{Zero-shot Summarization}

The main paper presents the zero-shot multilingual summarization experiments with GPT-4-EVAL \citep{gpt_eval_liu2023g-eval} as the main metric. In \Cref{table:xsum_rougel}, we present the same experiments with the more traditional ROUGE-L metric to provide more perspective and understanding of the results.

\begin{table}[h]
  \centering
  \resizebox{\columnwidth}{!}{%
  \setlength{\tabcolsep}{1.3pt}
  \begin{tabular}{l|c|c|c|c|c}
    \hline
            & {\bf Es} & {\bf Id} & {\bf Sw} & {\bf So} & {\bf Mr} \\
    \hline
    Basic	 & 12.7	& 12.8	& 12.2	& 11.5	&  4.1 \\
    XLT	     & 17.7	& 17.6	& 20.5	& 18.5	& 10.3 \\
    LDP	     & 18.1	& 18.6	& 21.8	& 19.0	& 10.0 \\
    LDP+U	 & \textbf{18.1} & \textbf{24.8} & \textbf{23.5}	& \textbf{19.3}	& \textbf{11.4} \\
    \hline
  \end{tabular}
  }
  \caption{
    ROUGE-L of different prompting techniques using InstructGPT text-davinci-003 for zero-shot summarization in high-resource (Es, Id) and low-resource (Sw, So, Mr) in the XL-sum summarization task \citep{xsum_Narayan2018DontGM}.
  }
  \label{table:xsum_rougel}
\end{table}

\end{document}